%% file: main.tex
\documentclass[runningheads]{llncs}
\usepackage{eccv}
\usepackage{eccvabbrv}

\usepackage{graphicx}
\usepackage{booktabs}

\usepackage[accsupp]{axessibility}  % Improves PDF readability for those with disabilities.

\usepackage[pagebackref,breaklinks,colorlinks,citecolor=eccvblue]{hyperref}
\usepackage{orcidlink}

\usepackage{colortbl}
\usepackage{xcolor}
\usepackage{colortbl}
\definecolor{colorh}{rgb}{1,0.60,0.20}
\definecolor{colorm}{rgb}{1,0.72,0.30}
\definecolor{colorl}{rgb}{1,0.88,0.70}
\newcommand{\colorh}[1]{\colorbox{colorh}{{#1}}}

\usepackage{graphicx}
\usepackage{multirow}
\usepackage{subcaption}

\usepackage[width=122mm,left=12mm,paperwidth=146mm,height=193mm,top=12mm,paperheight=217mm]{geometry}
\usepackage{bbding} % For \Envelope symbol
\usepackage{authblk}
\begin{document}

% ---------------------------------------------------------------
% TODO REVIEW: Replace with your title
\title{Multi-branch Collaborative Learning Network\\ for 3D Visual Grounding} 

% TODO REVIEW: If the paper title is too long for the running head, you can set
% an abbreviated paper title here. If not, comment out.
\titlerunning{Multi-branch Collaborative Learning Network for 3D Visual Grounding}

% TODO FINAL: Replace with your author list. 
% Include the authors' OCRID for the camera-ready version, if at all possible.
\author{
Zhipeng Qian\inst{*} \and
Yiwei Ma \inst{*} \and
Zhekai Lin \and
Jiayi Ji \and
Xiawu Zheng \and  \\
Xiaoshuai Sun~\Envelope   \and
Rongrong Ji }

% TODO FINAL: Replace with an abbreviated list of authors.
\authorrunning{Z.Qian et al.}
% First names are abbreviated in the running head.
% If there are more than two authors, 'et al.' is used.

% TODO FINAL: Replace with your institution list.
\institute{   Key Laboratory of Multimedia Trusted Perception and Efficient Computing, \\Ministry of Education of China, Xiamen University, Xiamen, Fujian, China.
\email{\{qianzhipeng,yiweima,linzhekai\}@stu.xmu.edu.cn},
\email{jjyxmu@gmail.com}, \\
% \email{yiweima@stu.xmu.edu.cn}, 
% \email{wanghaowei@stu.xmu.edu.cn}, 
\email{\{zhengxiawu,xssun,rrji\}@xmu.edu.cn}}

\maketitle
\renewcommand{\thefootnote}{\fnsymbol{footnote}}
\footnotetext{*Equal contributions.}
\footnotetext{\Envelope\ The corresponding author.}
\begin{abstract}
3D referring expression comprehension (3DREC) and segmentation (3DRES) have overlapping objectives, indicating their potential for collaboration. However, existing collaborative approaches predominantly depend on the results of one task to make predictions for the other, limiting effective collaboration.
We argue that employing separate branches for 3DREC and 3DRES tasks enhances the model's capacity to learn specific information for each task, enabling them to acquire complementary knowledge.
Thus, we propose the MCLN framework, which includes independent branches for 3DREC and 3DRES tasks. This enables dedicated exploration of each task and effective coordination between the branches.
Furthermore, to facilitate mutual reinforcement between these branches, we introduce a Relative Superpoint Aggregation (RSA) module and an Adaptive Soft Alignment (ASA) module. 
% The RSA module leverages the relative positional relationships among superpoints and their neighbouring points, enhancing the accuracy of superpoint feature aggregation through a more refined approach. The ASA module employs the query mask generated by the 3DREC branch to indirectly align the predicted bounding box and mask. It also improves the alignment ability of MCLN by adjusting Focal loss and Dice loss at the point and mask levels, respectively. 
These modules significantly contribute to the precise alignment of prediction results from the two branches, directing the module to allocate increased attention to key positions. 
% Finally, an adaptive weight is adopted to fuse the predictions and generate the final results. 
Comprehensive experimental evaluation demonstrates that our proposed method achieves state-of-the-art performance on both the 3DREC and 3DRES tasks, with an increase of \textbf{2.05\%}  in Acc@0.5 for 3DREC and \textbf{3.96\%} in mIoU for 3DRES. Our code is available at \url{https://github.com/qzp2018/MCLN}.
\end{abstract}

\begin{figure}[t]

    \centering
    \includegraphics[width=0.8\columnwidth]{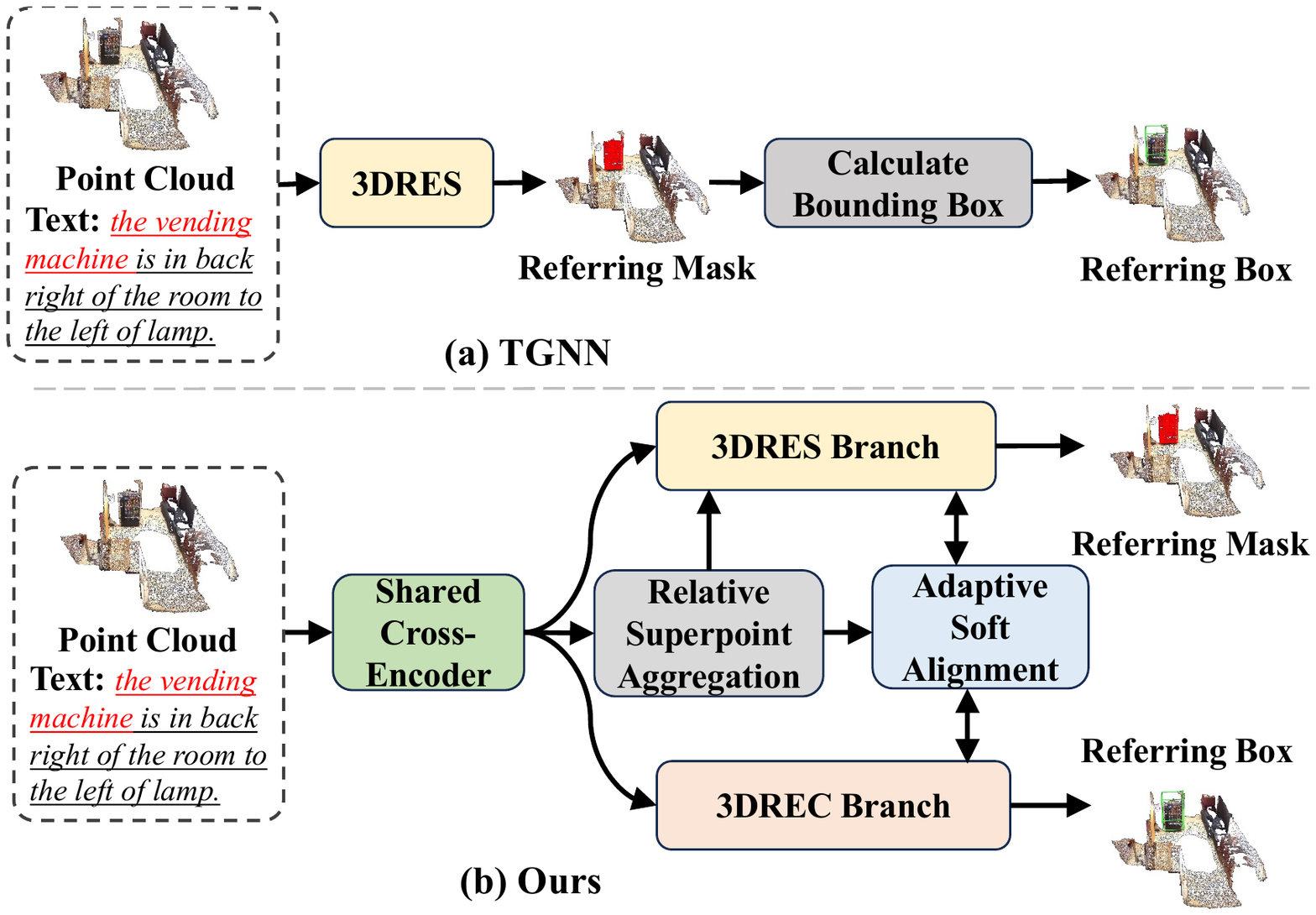}
    % \vspace{-1.5em}
    \caption
    {TGNN predicts the results of 3DREC and 3DRES tasks in a cascade manner. Our method introduces separate branches for 3DREC and 3DRES tasks and uses a parallel structure to facilitate collaborative training of them.}
%     {
%      (a) TGNN predicts the instance mask and obtains the bounding box using the maximum and minimum XYZ values within the segmentation mask. 
%      (b) Our method introduces separate branches for 3DREC and 3DRES tasks, along with a Relative Superpoint Aggregation (RSA) module and an Adaptive Soft
% Alignment (ASA) module to facilitate collaborative training of the two branches.
%       }
    \label{fig:intro}
% \vspace{-2.5em}
\end{figure}

\section{Introduction}
\label{sec:intro}
Understanding natural language and its relationship with visual information forms the fundamental basis for establishing a connection between humans and machines in the realm of artificial intelligence, and a 3D environment is often rich in visual information and meets realistic demands. 
Therefore, the task of 3D Visual Grounding (3DVG)~\cite{zhao20213dvg,yang2021sat,wu2022eda,chen2020scanrefer,cai20223djcg,Huang_Lee_Chen_Liu_2021} has gained increasing attention in recent years. Specifically, 3DVG involves precisely localising objects within a three-dimensional environment by interpreting natural language instructions.

The field of 3DVG involves two closely interconnected tasks: 3D Referring Expression Comprehension (3DREC)~\cite{wu2022eda,yuan2021instancerefer,zhao20213dvg,chen2020scanrefer} and 3D Referring Expression Segmentation (3DRES)~\cite{Huang_Lee_Chen_Liu_2021,2308.16632}. These tasks differ in how they ground targets. In 3DREC, targets are localized using bounding boxes, while in 3DRES, they are segmented into masks.
Traditionally, these tasks have been treated as separate entities despite their similar objectives.
Recently, attempts have been made to address both tasks simultaneously, such as the TGNN~\cite{Huang_Lee_Chen_Liu_2021} model. 
However, as depicted in Fig.~\ref{fig:intro} (a), TGNN's~\cite{Huang_Lee_Chen_Liu_2021} 3DREC performance relies entirely on the capabilities of the 3DRES, resulting in limited collaborative training benefits.

Hence, it is intuitive to explore the feasibility of incorporating distinct branches dedicated to predicting 3DREC and 3DRES outcomes while training them collaboratively. 
To solve this issue, we use superpoints to facilitate the joint training of 3DREC and 3DRES tasks. The Relative Superpoint Aggregation (RSA) module is proposed to effectively exploit the spatial relationships between superpoints and their surrounding points during the feature aggregation process. Such an approach could potentially mitigate individual shortcomings through mutual compensation. 
However, merely training these branches together is insufficient, as the separate branches introduce the risk of prediction conflicts, which could jeopardize the overall performance.
Therefore, ensuring alignment in the generated results from both branches is crucial.
Furthermore, achieving alignment between the branches is vital in facilitating the exchange of bidirectional information between 3DREC and 3DRES branches.
This collaborative learning approach empowers the branches to extract valuable information from one another, refining their positioning and contributing to a more effective performance.

Building upon the aforementioned analysis, we propose the incorporation of two distinct branches specifically designed for the 3DREC and 3DRES tasks.
By integrating separate branches, the model gains the capability to generate 3DREC and 3DRES results independently, preventing the degradation of one task due to dependency on the other.
Furthermore, it is natural to establish a direct correspondence between the bounding box generated by the 3DREC branch and the prediction mask generated by the 3DRES branch.
This can be achieved by leveraging superpoints within the bounding box.
However, it is important to note that the irregular shape of the superpoints and the potential inaccuracies in the predicted bounding boxes may lead to an incorrect representation of the grounding truth for those superpoints within the predicted bounding boxes. 
Therefore, a more rational and effective approach is to utilize the query mask generated from the 3DREC branch, which exhibits high consistency with the bounding box.
% After acquiring a potent indirect representation of the bounding box, the next step involves aligning the predicted bounding box with the mask generated from two distinct branches. 
After obtaining a robust indirect representation of the bounding box from the 3DREC branch, the next step involves aligning the predicted bounding box with the mask generated by the 3DRES branch.

To achieve this alignment, we propose the Adaptive Soft Alignment (ASA) module, which is a novel approach specifically designed to facilitate the simultaneous training of the 3DREC and 3DRES tasks. The ASA module enhances coordination between the two branches, leading to more effective and synchronized training.
Specifically, given that the query mask from the 3DREC branch may not always exhibit high quality or may contain inaccuracies in superpoint predictions, dynamically assigning weight coefficients during correspondence makes sense.
Hence, our proposed ASA module utilizes an adaptive approach to align the outcomes of the 3DREC and 3DRES tasks. 
It incorporates dynamic-weighted point-level Focal Loss and mask-level Dice Loss to enhance alignment within the 3DREC and 3DRES branches. 
The weights assigned to both losses are calculated dynamically, taking into consideration the quality of the query mask.
Lastly, an adaptive weight is applied to combine the predictions from the two branches and generate the final mask.

Our main contributions are summarized as follows:
\begin{itemize}
\item We introduce a novel multi-branch collaborative learning network, namely MCLN, for 3DVG, which comprises two distinct branches for collaborative learning between 3DREC and 3DRES tasks.

% This network comprises two distinct branches dedicated to addressing the 3DREC and 3DRES tasks independently. This design promotes collaborative learning between 3DREC and 3DRES tasks, alleviating the risk of one task's performance decline being solely reliant on the other.

\item To enhance the coordination and alignment between the 3DREC and 3DRES branches, we propose the integration of a Relative Superpoint
Aggregation (RSA) module and an Adaptive Soft Alignment (ASA) module. 
% This module integrates two meticulously crafted losses: a mask-level Dice loss and a point-level Focal loss designed to enhance correspondence. Additionally, we utilize a query mask generated by the 3DREC branch to establish the correspondence, along with an adaptive weight to effectively merge the masks produced by two branches.

\item Our proposed method has achieved state-of-the-art performance in both 3DREC and 3DRES tasks, showcasing a notable improvement of $\textbf{2.05\%}$ Acc@0.5 for 3DREC and a substantial  increase of $\textbf{3.96\%}$ mIoU for 3DRES.
\end{itemize}

\section{Related Work}
\subsection{3D Visual Grouding}
3D vision and language play pivotal roles in human comprehension of the world, and they serve as significant areas of research for advancing the development of machines with human-like capabilities. Consequently, noteworthy research endeavours in vision and language~\cite{yang2024sam,9761944,9802801,10.1145/3394171.3414009,ma2022xclip,ma2023xmesh,ma2023towards} have surged. Among these, 3D Visual Grounding (3DVG)~\cite{zhao20213dvg,yang2021sat,wu2022eda, Huang_Lee_Chen_Liu_2021,chen2020scanrefer,cai20223djcg,lin2023unified,chen2022language} stands out as a prime exemplar, focusing on the task of precisely localizing objects referred to in natural language descriptions. In a comprehensive scope, 3DVG comprises two fundamental tasks: 3D Referring Expression
Comprehension (3DREC)~\cite{zhao20213dvg,yang2021sat,wu2022eda,chen2020scanrefer,cai20223djcg,yuan2021instancerefer,chen2022language} and 3D Referring Expression
Segmentation (3DRES)~\cite{Huang_Lee_Chen_Liu_2021,2308.16632,qian2024x}. For 3DREC, most of the existing works~\cite{chen2020scanrefer,yang2021sat,yuan2021instancerefer,cai20223djcg} take a two-stage approach. Initially, candidate proposals are obtained using a pre-trained 3D object detector or ground truth information. Subsequently, visual and text features are extracted for multi-modal feature fusion, and then the most suitable proposal is selected based on the fused features. Recently, single-stage methods~\cite{wu2022eda,jain2022bottom} have also shown superior performance, presenting another promising paradigm. Conversely, although the 3DREC task has received considerable attention, the 3DRES task was initially introduced by TGNN~\cite{Huang_Lee_Chen_Liu_2021}, with limited exploration in subsequent years. Recently there has been a flurry of work to investigate this area. For example, 3D-STMN~\cite{2308.16632} employs superpoints for the 3DRES task, greatly improving performance. However, the collaborative work between 3DREC and 3DRES is still very few, indicating significant untapped potential.

\subsection{Multi-task Learning}
Multi-task learning (MTL)~\cite{Luo_Zhou_Sun_Cao_Wu_Deng_Ji_2020, Caruana_Pratt_Thrun_2017, Li_Zhang_Sun_Wu_Zhao_Tan_2022, Nekrasov_Dharmasiri_Spek_Drummond_Shen_Reid_2018, Hua_Liao_Tian_Zhang_Zou_2023} serves as a valuable strategy for enhancing task performance, effectively leveraging training data while facilitating the exchange of information among related tasks. MTL has found extensive applications across a diverse range of computer vision tasks. In the 2D vision, many characteristics are shared between detection and segmentation, which have prompted several works~\cite{He_Gkioxari_Dollar_Girshick_2020, Bolya_Zhou_Xiao_Lee_2019, Fu_Shvets_Berg_2019,lin2023unified} to address both tasks within a unified framework. Similarly, REC and RES can also promote each other in joint training~\cite{Luo_Zhou_Sun_Cao_Wu_Deng_Ji_2020, Hua_Liao_Tian_Zhang_Zou_2023, Li_Zhang_Sun_Wu_Zhao_Tan_2022}. However, due to the limited exploration of 3DRES in recent years, collaborative initiatives between 3DRES and 3DREC have mainly remained uncharted.  

TGNN~\cite{Huang_Lee_Chen_Liu_2021} conducts the first attempt to employ a single model to tackle both 3DREC and 3DRES tasks. However, it doesn't set up a separate branch for the 3DREC task, and this design exhibits several limitations. Specifically, it ties the performance ceiling for 3DREC solely to 3DRES, overlooking the distinctions between these tasks. Similarly, 3DRefTR~\cite{lin2023unified} also adopts the cascade structure to tackle two tasks, severely restricting the interaction between the 3DREC and 3DRES tasks and leading to sub-optimal performance. To deal with the above problems, we designed two separate branches for 3DREC and 3DRES and a co-alignment module for joint training.

\section{Method}
\begin{figure*}[t]
\centering 
\includegraphics[width=1\columnwidth]{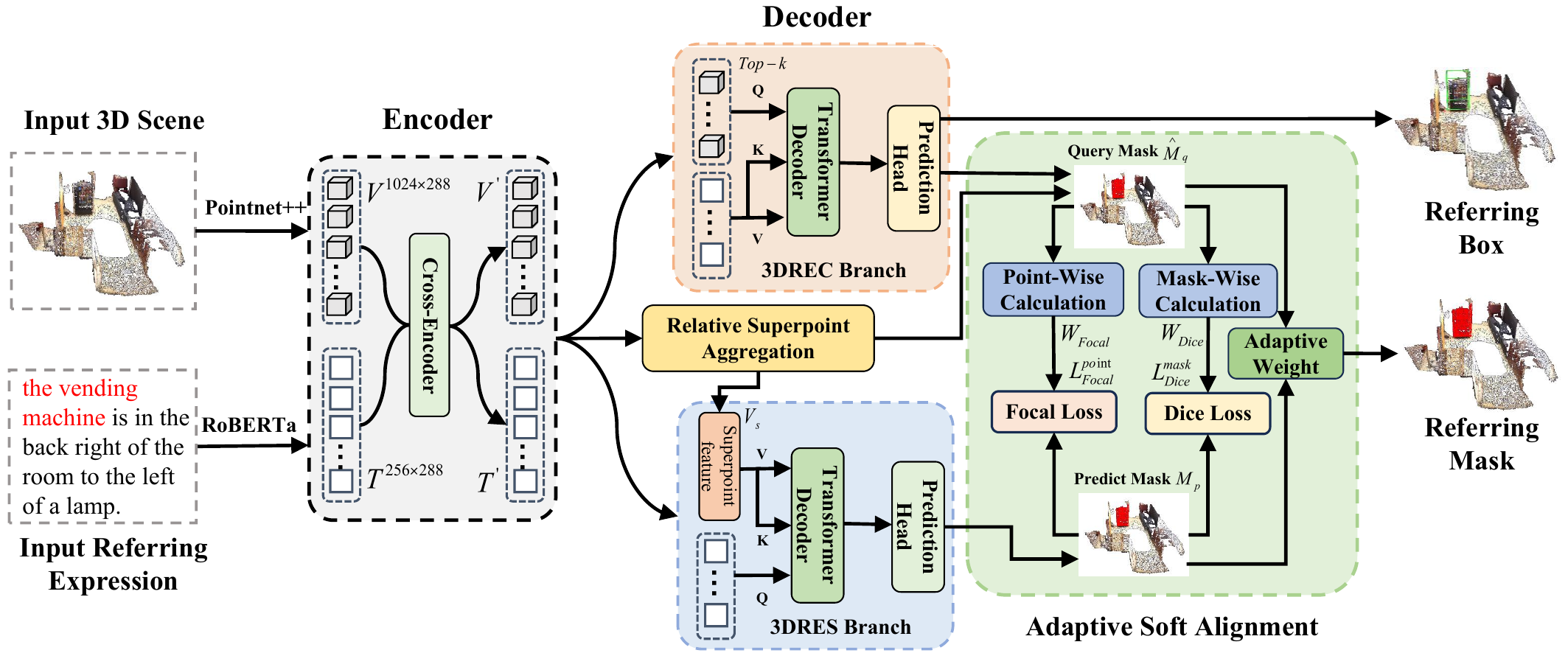}
  % \vspace{-2em}
  \caption{An overview of our proposed network. Our proposed network comprises a cross-modal encoder, separate decoders for 3DREC and 3DRES, a “Relative Superpoint Aggregation” (RSA) module for superpoint feature generation, and an “Adaptive Soft Alignment” (ASA) module for joint training. 
  % The RSA module utilizes relative position relationships for feature aggregation, while the ASA module includes “Soft Alignment” with point-level and mask-level weight adjustments for Dice and Focal losses, and “Adaptive Weight” for dynamic fusion of predicted masks from two branches.
  }
  \label{fig:overview}
  % \vspace{-2em}
\end{figure*}

The framework of the proposed Multi-branch Collaborative Learning Network (MCLN) is shown in Fig.~\ref{fig:overview}. Since there are numerous existing 3DREC models with superior performance, we adopt a typical DETR-like model EDA~\cite{wu2022eda}, which is composed of a backbone, a cross-encoder, and a 3DREC decoder. Then we designed a segmentation head for the task of 3DRES, which shares the backbone and the cross-modal encoder with the 3DREC branch. We also propose a “Relative Superpoint Aggregation” (RSA) module to generate superpoint features by exploiting the relative spatial relationships among nearby visual points. This module enables 3DREC and 3DRES branches to share similar visual features, making further alignment possible. In addition, an “Adaptive Soft Alignment” (ASA) module is proposed to maximize the consistency between the results of 3DREC and 3DRES branches and promote collaborative training between them. 

\subsection{The Framework}
As shown in Fig.~\ref{fig:overview}, branches of 3DRES and 3DREC share the same backbone and the cross-modal encoder, where the inference heads remain relatively independent. Specifically, following EDA~\cite{wu2022eda}, we adopt the pre-trained RoBERTa~\cite{Liu_Ott_Goyal_Du_Joshi_Chen_Levy_Lewis_Zettlemoyer_Stoyanov} and PointNet++~\cite{Qi_Yi_Su_Guibas_2017} to produce text tokens $ \textit{T} \in \mathbb{R}^{l\times d}$ and visual tokens $\textit{V} \in \mathbb{R}^{n\times d}$, where $l=256$ and $n=1024$ represent the numbers of text tokens and visual tokens, $d=288$ represents the feature dimension. Then the visual tokens $\textit{V}$ and text tokens $\textit{T}$ are put into the cross-modal encoder, where self-attention and cross-attention are performed to update both visual and text features, obtaining cross-modal features $\textit{V}' \in \mathbb{R}^{n\times d}$ and $\textit{T}'\in \mathbb{R}^{l\times d}$.

\subsection{Multi branches for 3DREC and 3DRES }
Given that the objectives of both tasks remain distinct, complete sharing of the inference branch may prove counterproductive. So such two independent branches along with an alignment module can help models better adapt to their respective tasks and have more independent and valid information. 

\subsubsection{3DREC Bounding Box Prediction.}
For the 3DREC branch, we follow the structure of EDA~\cite{wu2022eda}, the top-k (k=256) object queries $\textit{O}\in \mathbb{R}^{k\times d}$ produced by the cross-modal encoder are obtained and put into the 3DREC decoder, and position alignment loss and semantic alignment loss are designed respectively for box regression and box-text alignment. Given the predicted bounding boxes  $\textit{B}\in \mathbb{R}^{k\times 6}$ and alignment score $\textbf{sc}\in \mathbb{R}^{k}$, the 3DREC branch could localize the referred object by $\textit{B}_{box}\in \mathbb{R}^{1\times 6}$ with highest score among \textbf{sc}. It's recommended to refer to EDA~\cite{wu2022eda} for more details.
% For the 3DRES branch, taking inspiration from SPFormer~\cite{Sun_Qing_Tan_Xu_2022} and 3D-STMN~\cite{2308.16632}, our design incorporates a sequence of transformer decoder layers. Similarly, we also adopt superpoints for 3DRES, which serve as a bridge connecting low-resolution features to the high-resolution point cloud, facilitating a more efficient upsampling technique. 
\subsubsection{Relative Superpoint Aggregation.}
For the 3DRES task, dense point clouds are essential for the representation of masks. However, the Pointnet++~\cite{Qi_Yi_Su_Guibas_2017} 3D feature backbone filters out only a few point cloud features, making it difficult to directly project them back into the segmentation masks due to the sparsity of the remaining point clouds.
To solve this issue, taking inspiration from previous methods~\cite{Sun_Qing_Tan_Xu_2022,2308.16632,han2020occuseg,liang2021instance,lin2023unified}, we adopt superpoints, which are collections of points with specific attributes or characteristics in point cloud data. Superpoints serve as a bridge connecting low-resolution features to the high-resolution point cloud, making the joint training of 3DREC and 3DRES tasks possible. 
To obtain the superpoint-level features $\textit{V}_s\in \mathbb{R}^{m \times d}$ ($m$ and $d$ represent the number of superpoints and feature dimensions respectively), we make use of the refined visual point features $\textit{V}'$ produced by the cross-modal encoder before. Specifically, we utilize the ball query method~\cite{Qi_Yi_Su_Guibas_2017} to identify the $K$ nearby visual point features within $R$ radius of coordinate space for each superpoint (we set $K=2, R=0.2$). However, since there are still existing position deviations between the superpoints' centres and their nearby points, directly using the ball query ignores the relative position relation between superpoints and nearby points, leading to inaccurate superpoint features aggregation. Taking the consideration into account, we denote the centre of $s$-th superpoint as $o_s$, its feature as $v_s$, the centres of the $K$ nearby points as $N_s=\{o_s^1,o_s^2,...,o_s^K\}$, then we use an MLP layer to encode the relative coordinate of the $k$-th neighbour of superpoint $s$:
\begin{equation}
    r_s^k=\mathrm{MLP}(o_s-o_s^k).
\end{equation}
Then we add the relative coordinate features to the nearby $K$ visual point features $\{v_s^1,v_s^2,...,v_s^K\}$ and employ max-pooling for aggregation, producing the $s$-th superpoint feature $\textit{v}_s\in \mathbb{R}^{1 \times d}$ ( $\textit{v}_s\in \textit{V}_s$).
\begin{eqnarray}
    \hat{v}_s^k&=&r_s^k+v_s^k,(k\in [1,K])\\
    v_s&=&\mathrm{Maxpool}(\hat{v}_s^1,\hat{v}_s^2,...,\hat{v}_s^K).
\end{eqnarray}
\subsubsection{3DRES Mask Prediction.}
For the 3DRES branch, our design incorporates a sequence of transformer decoder layers.
After superpoint features generation, textual query vectors fully integrate scene information via  making cross-attention with superpoints, which can be formulated as:
\begin{equation}
   S_l=\mathrm{softmax}(\frac{QK^T}{\sqrt{D}}+A_{\ell-1})V,
\end{equation}
where $ S_{\ell}\in \mathbb{R}^{l\times d}$ is the refined textual query vectors (notably ${S_0}=\textit{V}'$), $Q=\phi_Q(S_{\ell-1})\in \mathbb{R}^{l \times d}$ is obtained through the linear projection $\phi_Q$ of the text query $S_{\ell-1}$, and $K$,$V$ is obtained through the linear projection $\phi_K$,  $\phi_V$ of the superpoint features $\textit{V}_s$. $A_{\ell-1}\in \mathbb{R}^{l \times m}$ is the superpoint attention mask, and $l$ and $m$ is the number of textual tokens and superpoints. To get $A_{\ell-1}$, we first get the superpoint attention map $M_{\ell-1}\in \mathbb{R}^{l \times m}$ by computing the similarity between textual query $S_{\ell-1}$ and the superpoint features  $\textit{V}_s$, then filter superpoints with a threshold $\tau$, as:
\begin{eqnarray}
    M_{\ell-1}&=&S_{\ell-1}\times V_s,\\
    A_{\ell-1}{(i,j)} &=&
    \begin{cases}
    0 & \text{if } \mathrm{Sigmoid}(M_{\ell-1}{(i,j)}) \geq \tau \\
    -\infty & \text{otherwise}
    \end{cases}.
\end{eqnarray}
${A_{\ell-1}{(i,j)}}$ indicates $i$-th word token attending to $j$-th superpoint if ${M_{\ell-1}{(i,j)}}$ is higher than $\tau$. Empirically, we set $\tau$ to 0.5.
By stacking transformer decoder layers, the superpoint attention mask $A_{\ell-1}$ adaptively restricts cross-attention within the target superpoints.

Following multiple iterations of updating the textual query, we get the refined textual tokens $S_l$. Subsequently, by summing the similarity scores between textual tokens $S_l$ and all the superpoint features $V_s$, we select $S_{h}$, the textual token with the highest score. Then we forward $S_{h}$ to the prediction head, by computing the similarity between $S_{h}$ and $V_s$,  we derive the predicted mask $M_{p}\in \mathbb{R}^{1 \times m}$:
\begin{eqnarray}
    M_{p}&=&S_{h}\times{V_{s}}, 
\end{eqnarray}
where  $S_{h}\in \mathbb{R}^{1 \times d}$ and superpoint features $V_s\in \mathbb{R}^{m \times d}$. 

\subsection{Adaptive Soft Alignment }
To promote collaborative training between the two branches, we further propose a module called \textit{Adaptive Soft Alignment} (ASA) to mitigate the influence of prediction inconsistency by learning the inter-task relationships. Specifically, the module consists of ``Soft Alignment” and ``Adaptive Alignment” two parts.

For ``Soft Alignment”, since we have obtained the predicted superpoint mask $M_p$ from the 3DRES branch, we need to make alignment to the output bounding box of the 3DREC branch. However, due to the discrepancy in the presentation format of results between the two tasks, directly aligning the output mask with the bounding box may result in inaccuracies and yield suboptimal performance. 
To solve the issue above, we introduce the query mask $\hat{M}_{q}\in \mathbb{R}^{1 \times m}$, obtained by selecting superpoints with high similarity to the query vector $Q_{box}\in \mathbb{R}^{1 \times d}$, which was used to predict the bounding box in the 3DREC branch:
\begin{eqnarray}
    M_q&=&Q_{box}\times V_s,\\
    \hat{M}_q^i&=
    &\begin{cases}
    1 & \text{if} \ {M}_q^i \geq \tau \\
    0 & \text{otherwise}
    \end{cases},
\end{eqnarray}
where  $\hat{M}_q^i$ represents the $i$-th value of $\hat{M}_q$.

The query mask $\hat{M}_{q}$ acts as an indirect representation of the predicted bounding box. It serves as a connecting link, facilitating the alignment between the bounding box from the 3DREC branch and the mask from the 3DRES branch. This soft alignment scheme enhances the coherence between the two branches.
% Specifically, within the 3DREC branch, the selected query used to generate the bounding box produces a superpoint mask prediction denoted as ${M}_{q}$. This mask ${M}_{q}$ is generated by conducting a similarity computation between the superpoint features $V_s$ and the chosen query embeddings $Q$ within the 3DREC branch, as detailed in work ~\cite{lin2023unified}. Consequently, ${M}_{q}$ exhibits significant consistency with the bounding box prediction, as both are derived from the same query embedding. 
Through the alignment of superpoint masks ${M}_{p}$ and $
\hat{M}_{q}$, the indirect preservation of spatial coherence between the outcomes of the 3DREC and 3DRES tasks is effectively achieved. 
Usually, a Focal loss~\cite{Lin_Goyal_Girshick_He_Dollar_2017} and a Dice loss~\cite{Milletari_Navab_Ahmadi_2016} will be applied to conduct the alignment between masks.
% {\small
% \begin{equation}
% \begin{aligned}
% L_{\text{Focal}}(M_p, \hat{M}_q) =  \sum_{i=1}^{m}-\left( \hat{M}_q^i \cdot (1 - M_p^i)^\gamma \cdot \log(M_p^i) \right.\\
%  \left. + (1 - \hat{M}_q^i) \cdot (M_p^i)^\gamma \cdot \log(1 - M_p^i)\right),
% \end{aligned}
% \end{equation}
% }
% \begin{eqnarray}
% L_{\text{Dice}}(M_p, \hat{M}_q) = \sum_{i=1}^{m} \left(1 - \frac{2|M_p^i \cap \hat{M}_q^i|}{|M_p^i| + |\hat{M}_q^i|}\right),
% \end{eqnarray}
% where $m$ represents the number of superpoints, $M_p^i$ and $\hat{M}_q^i$ represents the $i$-th value of $M_p$ and $\hat{M}_q$, and $\gamma=2$.

However, inspired by ~\cite{Yang_Ji_Sun_Wang_Li_Ma_Ji_2023}, we argue that the quality of the query mask $M_q$ generated by the REC branch, utilized for alignment, is also a crucial factor in the alignment process. So we adopt two different quality assessment methods tailored to these two distinct loss functions, referred to as the “Adaptive Alignment”.
\subsubsection{Point-wise Focal Loss.}
As the Focal loss is computed on a per-superpoint basis, it is categorized as a point-level loss. In essence, when the value of the element in the output query mask $
{M}_{q}$ of the REC branch is close to 0 or 1, and the superpoint is more confidently identified as either background or foreground, leading to a higher-quality label for that superpoint. Conversely, when the probability approaches 0.5, the model's prediction for that superpoint becomes more ambiguous. Consequently, in the calculation of Focal loss, we assert that all superpoints should not be treated equally. Instead, higher-quality superpoints' labels should be assigned higher weights to reflect their increased certainty. Therefore, we design a normal distribution algorithm formula to measure the quality of each superpoint in the query mask $M_q \in \mathbb{R}^{1 \times m}$:
\begin{equation}
    W_\text{Focal}({M}_{q}^{i}) =b-\frac{1}{\sqrt{2\pi}\sigma}exp\left(-\frac{({{M}_{q}^{i}}-\mu)^2}{2\sigma^2}\right),
\end{equation}
where ${M}_{q}^{i}$ represents the $i$-th value of the query mask ${M}_{q}$ from 3DREC branch. We set $b=2,\mu=0.5,\sigma^2=0.1$ to limit the scoring range roughly to $(0.7,2)$. Lastly, the Focal loss can be rewritten as:
\begin{equation}
L_\text{Focal}^{point}({M}_{p},\hat{M}_{q})=\sum_{i=1}^{m} W_\text{Focal}({M}_{q}^{i}) * L_\text{Focal}({M}_{p}^{i},\hat{M}_{q}^i).
\end{equation}

\subsubsection{Mask-wise Dice Loss.}
Unlike Focal loss, Dice loss treats the predicted and ground truth masks holistically, emphasizing the overlapping regions between them. This attribute of Dice loss renders it particularly well-suited as a mask-level loss. We argue that the average distance between the selected superpoints is a good basis for measuring the performance of the mask. In the case of an accurately predicted mask, its superpoints are consistently close to each other, whereas incorrectly predicted query masks may exhibit dispersed superpoints. So the quality of query masks can be assessed through the calculation of mean distances among the superpoints' centres within them:
\begin{equation}
    W_\text{Dice}(\hat{M}_{q}) =\frac{1}{1+\sum_{\substack{i,j \\ i \neq j}} \frac{2d_{i,j}}{N*(N-1)}},    
               \quad \text{where} \ \hat{M}_{q}^i,\hat{M}_{q}^j==1,
\end{equation}
where $\hat{M}_{q}^i$ and $\hat{M}_{q}^j$ means the $i$-th and $j$-th value of Query Mask $\hat{M}_{q}$, $d_{i,j}$ means the distance between $i$-th and $j$-th superpoints. $N$ is the number of superpoints whose values are $1$.
Finally, the Dice loss is reformulated as:
\begin{equation}
L_\text{Dice}^{mask}(M_p, \hat{M}_q) = W_\text{Dice}(\hat{M}_{q}) * L_\text{Dice}({M}_{p},\hat{M}_{q}).
\end{equation}

% \begin{figure}[] % 使用figure*使图片横跨双栏宽度
%   \centering 
%   \includegraphics[width=\columnwidth]{picture/good_bad_cases.pdf} % 使用\textwidth确保图片横跨双栏宽度
%   \vspace{-3em}
%   \caption{Example of low-quality and high-quality query masks, The low-quality query mask has scattered superpoints, while the high-quality query mask's superpoints are clustered. So the mean distance between superpoints of the low-quality query mask is much larger than the high-quality query mask.}
%   \vspace{-1em}
%   \label{fig:goog_bad_cases}
% \end{figure}

While alignment through the loss function is employed, the coordination between the two branches may still have limitations. To overcome this, we incorporate an additional confidence prediction within the 3DRES branch to evaluate the credibility of the predicted mask ${M}_{p}$. Subsequently, we merge the two predicted superpoint masks ${M}_{p}$ and ${M}_{q}$ based on the confidence scores. Through the adaptive assembling of the two superpoint masks, we integrate information from both sources, culminating in the generation of the final superpoint mask, denoted as $M_{f} \in \mathbb{R}^{1 \times m}$.
\begin{eqnarray}
    \mu&=&\mathrm{Sigmoid(MLP}(S_{h})),\\
    M_{f}&=&\mu*{M}_{p}+(1-\mu)*{M}_{q},
\end{eqnarray}
where $S_{h}\in \mathbb{R}^{1 \times d}$ represents the textual token with the highest score, and $\mu$ is the confidence score.    During inference, the final predicted mask is obtained by filtering superpoints with the threshold $\tau=0.5$.
\subsection{Overall Loss}
For the 3DREC task, we use a training approach proposed by EDA~\cite{wu2022eda}. It involves five loss functions for each layer of the object decoder: smooth-L1 loss for box centre coordinate prediction ($L_\text{coord}$) and box size prediction ($L_\text{size}$), GIoU loss~\cite{Rezatofighi_Tsoi_Gwak_Sadeghian_Reid_Savarese_2019} ($L_\text{giou}$), semantic alignment loss ($L_\text{sem}$), and position alignment loss ($L_\text{pos}$). The loss for the $i$-th decoder layer is the result of combining these five terms through weighted summation. 
\begin{equation}
    L_\text{dec}^{i}=\alpha_1 L_\text{coord}^{i}+\alpha_2 L_\text{size}^{i}+\alpha_3 L_\text{giou}^{i}+\alpha_4 L_\text{sem}^{i}+\alpha_5 L_\text{pos}^{i}. 
\end{equation}

The losses across all decoder layers are averaged by the number of layers $L$ to compute the overall 3DREC loss:
\begin{equation}
    L_{rec}=\frac{1}{L}\sum_{i=1}^{L} L_{dec}^{i}.
\end{equation}

For the 3DRES task, we adopt the Focal loss and the Dice loss for two branches' masks and alignment loss:
\begin{equation}
    L_\text{res}=\beta_1( L_\text{Focal}^1+ L_\text{Focal}^2+L_\text{Focal}^{fin}+ L_\text{Focal}^{point})+\beta_2( L_\text{Dice}^1+ L_\text{Dice}^2+ L_\text{Dice}^{fin}+ L_\text{Dice}^{mask}),
\end{equation}
where $L_\text{Focal}^1, L_\text{Dice}^1$ denotes the losses associated with the superpoint mask ${M}_{f}$ and the ground truth. Similarly, $L_\text{Focal}^2, L_\text{Dice}^2$ represents the losses for the superpoint mask ${M}_{q}$. Additionally, $L_\text{Focal}^{fin}, L_\text{Dice}^{fin}$ correspond to the losses for the final predicted superpoint mask ${M}_{f}$. Finally, $L_\text{Focal}^{point}$ and $L_\text{Dice}^{mask}$ denote the adaptive alignment losses between ${M}_{p}$ and $\hat{M}_{q}$.

Together with the KPS loss~\cite{Liu_Zhang_Cao_Hu_Tong_2021}, the final loss is as follows:
\begin{equation}
   L = \gamma_1 L_\text{rec} + \gamma_2 L_\text{res} + \gamma_3 L_\text{kps} .
\end{equation}

\section{Experiments}

% \subsection{Dataset}
We evaluate our method on 3D referring datasets, such as ScanRefer~\cite{chen2020scanrefer}. ScanRefer dataset consists of 1,513 intricately reconstructed 3D indoor scenes, following the official ScanNet~\cite{Dai_Chang_Savva_Halber_Funkhouser_Niessner_2017} splits. 
% ScanRefer consists of 51,583 descriptions corresponding to 11,046 3D objects extracted from 800 real-world scenes captured with the ScanNet dataset. This dataset is notable for being the first extensive collection designed to support 3D object grounding in point clouds using intricate and diverse natural language descriptions.
Additional experiments on \textbf{SR3D/NR3D} datasets and more qualitative results are detailed in the Appendix.
% On average, each scene comprises 13.81 objects and 64.48 descriptions, while each object is linked to an average of 4.67 descriptions. This abundance of data offers a rich and diverse resource for tasks related to associating objects with textual references.
\input{table/REC_table}

\subsection{Implementation Details}
Our model undergoes training using the AdamW optimizer with a batch size of 12, leveraging the computational power of four 24-GB NVIDIA RTX-3090 GPUs. During training, the text backbone Robeta-base is frozen, while the remaining components of the network remain trainable. Regarding the ScanRefer dataset, we employ a learning rate of $2e-3$ for the visual encoder and $2e-4$ for all other layers. To address balancing factors, we set $\tau=0.5$, $L=6$, $\alpha_1=5$, $\alpha_2=1$, $\alpha_3=1$, $\alpha_4=0.5$, $\alpha_5=0.5$, $\beta_1=10$, $\beta_2=2$, $\gamma_1=1/(L+1)$, $\gamma_2=1$ and $\gamma_3=8$.

\input{table/RES_table}
\subsection{Quantitative comparisons}
In this subsection, experiments on ScanRefer show that our method achieves SOTA performance in both 3DREC and 3DRES tasks.

\noindent\textbf{3DREC.} We adopt Acc@0.25,Acc@0.5 as indicators to evaluate the 3DREC performance. As shown in Tab.~\ref{tab:REC}, our method consistently exhibits superior performance over these prevailing methods in both two-stage and single-stage methods. Notably, in the realm of two-stage methods, our approach outperforms others in all performance metrics by a large margin (\textbf{2.05\%} in Acc@0.5). This evidence strongly suggests that the utilization of two separate branches facilitates the learning of task-specific information, leading to a substantial improvement in the performance of the 3DREC task through collaborative training between the branches. Furthermore, the remarkable results can be largely attributed to the well-designed Relative Superpoint Aggregation (RSA) module and Adaptive Soft Alignment (ASA) module, underscoring their effectiveness in the overall performance of the model. Nevertheless, in the single-stage approach, there is a decrease in performance under unique scenes. This could be attributed to the heightened recognition capability in complex scenes, potentially resulting in a lack of adaptability to simpler scenes.

\noindent\textbf{3DRES.} We utilize the evaluation metrics proposed earlier: {mean IoU} and {Acc@kIoU}. As demonstrated in Tab.~\ref{tab:RES}, our method showcases a substantial performance improvement when compared to the previous method TGNN~\cite{Huang_Lee_Chen_Liu_2021}, as well as the latest approaches~\cite{qian2024x,2308.16632,lin2023unified}. Notably, our method achieves impressive increases of \textbf{4.63\%} in Acc@0.5, and \textbf{3.96\%} in mIoU. This highlights the efficacy of aligning the two branches, enabling the model to distinguish and comprehend relationships in complex scenes.

\input{table/Alignment_and_mulplus}

\input{table/RSA-ASA_and_Focal-Dice}
\subsection{Ablation Study}
\subsubsection{Alignment Targets Ablation.}
We performed an ablation study to evaluate the impact of various target alignment strategies. The “Mask” refers to the alignment between query mask ${M}_{q}\in \mathbb{R}^{1 \times m}$ and predict mask ${M}_{p}\in \mathbb{R}^{1 \times m}$. While the “Box” denotes the direct alignment between the predicted bounding box $\textit{B}_{box}\in \mathbb{R}^{1\times 6}$ and ${M}_{p}\in \mathbb{R}^{1 \times m}$, which is realized by making alignment between predicted mask ${M}_{p}$ and superpoints whose centres falling within the bounding box $\textit{B}_{box}$. As depicted in Tab.~\ref{Alignment}, alignment targets derived from the query mask consistently outperform alignment using bounding boxes across all metrics. This observation emphasizes that direct alignment based on the bounding box may lead to omissions or inaccuracies, ultimately resulting in sub-optimal outcomes.

\subsubsection{Fusion methods Ablation.}
In the Relative Superpoint Aggregation (RSA) module, we try two different methods to fuse relative position vectors with point cloud features: “Plus” is shown in Equ.~2, “Mul”  refers to the process of multiplying relative position vectors with point cloud features, followed by adding the fused features to the original point cloud features. As shown in Tab.~\ref{Fusion}, “Plus” surpasses “Mul”, especially in 3DREC metrics. This may be because the “Mul” fusion method induces substantial alterations to point features. As a result, the superpoint features obtained through the “Mul” fusion differ significantly from the original point cloud features. Consequently, during the alignment of the two modules, the outputs from the two branches don't align well, leading to suboptimal performance. The “Plus” fusion method not only incorporates the relative spatial information of superpoints but also endeavours to preserve the original features of the points to a significant extent.

\subsubsection{Module Ablation.}
We performed an ablation experiment to investigate the impact of the Adaptive Soft Alignment (ASA) module and the Relative Superpoint Aggregation (RSA) module. The results, as depicted in Tab.~\ref{ASA}, reveal that the basic MCLN, initially equipped with two branches for the 3DREC and 3DRES tasks and simply using ball query for Superpoint Aggregation, has already surpassed the performance of 3D-STMN~\cite{2308.16632} and EDA~\cite{wu2022eda}. Notably, upon the integration of the ASA module, our model exhibits a substantial performance boost in all metrics. This suggests that our ASA module effectively facilitates the two branches in learning complementary information and enhances their coordination during training. Moreover, the ASA module results in a notable improvement in our model's performance, particularly in the 3DRES task. This demonstrates that our RSA module contributes to a more precise aggregation of superpoint features, underscoring the efficacy of the RSA module. Additionally, the ASA module's effectiveness in alignment is also substantiated.

\subsubsection{Ablation study of Adaptive Soft Alignment approach.} 
In Tab.~\ref{Ablation_Point_Wise}, we initially assess the efficacy of our proposed Quality-Based loss adjustment approach, involving point-wise weights for Focal loss and mask-wise weights for Dice loss. As illustrated in the table, both point-wise Focal loss and mask-wise Dice loss contribute significantly to improvements across all metrics, proving their effectiveness. Notably, the amalgamation of point-wise Focal loss and mask-wise Dice loss further improves the model performance and attains state-of-the-art (SOTA) performance in both 3DREC and 3DRES tasks, demonstrating a certain degree of complementarity between these two loss functions.
% Specifically, this combination leads to a substantial increase in Acc@0.5 for the 3DREC task, accompanied by the SOTA performance of mIoU in the 3DRES task. 
% Nevertheless, when compared to the baseline, our model exhibits a slight decrease in Acc0.25 for the 3DRES task. We posit that the reduction in Acc0.25 may be attributed to a corresponding enhancement in Acc0.5.

\begin{figure*}[t] % 使用figure*使图片横跨双栏宽度
  \centering 
  \includegraphics[width=1.0 \textwidth]{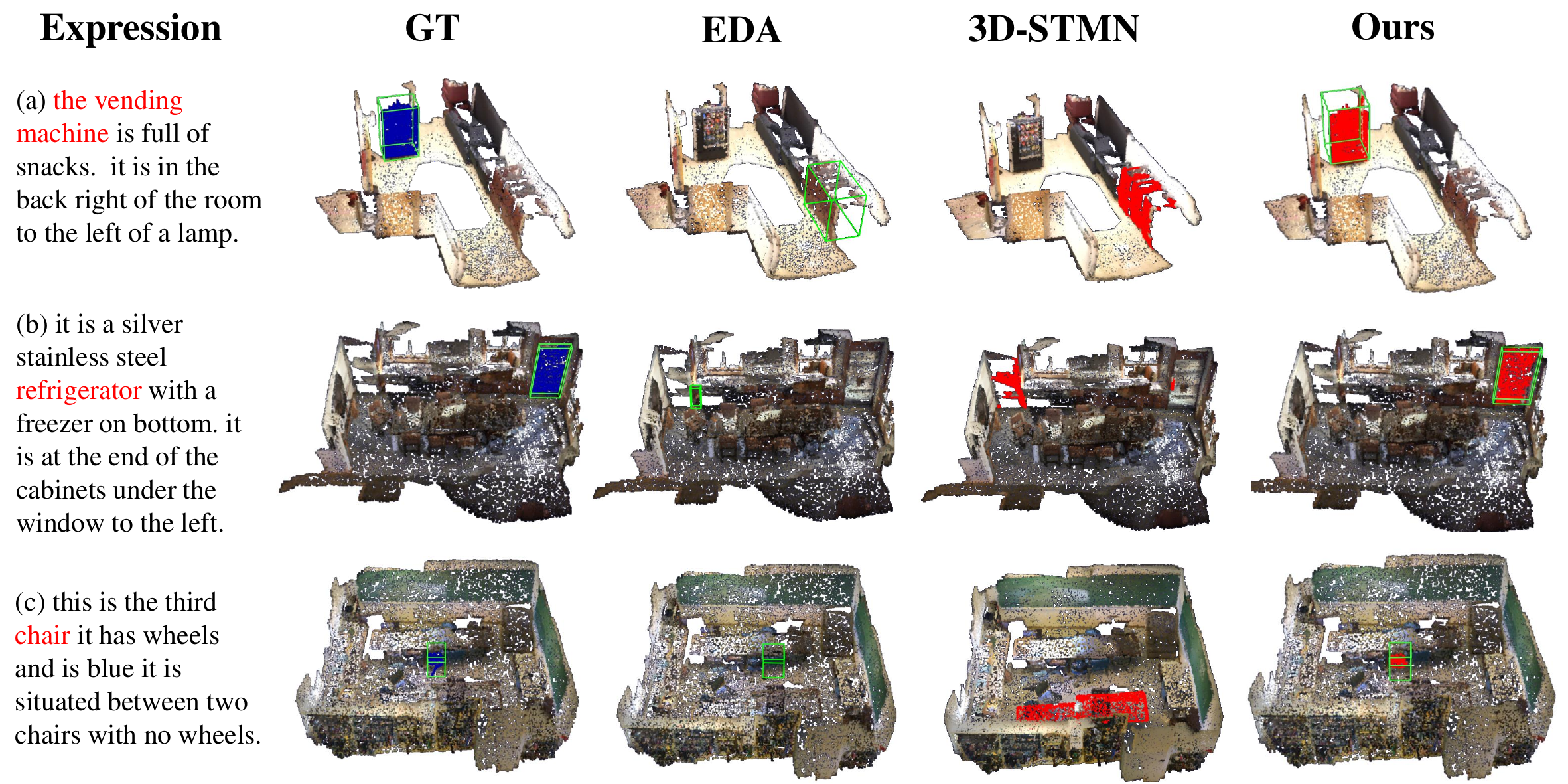} % 使用\textwidth确保图片横跨双栏宽度
  % \vspace{-2em}
  \caption{Qualitative results from 3D-STMN~\cite{2308.16632}, EDA ~\cite{wu2022eda} and our model.}
  % Our model exhibits superior positioning ability compared to 3D-STMN~\cite{2308.16632} and EDA ~\cite{wu2022eda}, along with a high level of consistency between the referred box and mask.}
  \label{fig:keshihua}
  % \vspace{-2em}
  
\end{figure*}

\subsection{Visualization}
Fig.~\ref{fig:keshihua} shows three qualitative referring detection and segment results produced by the EDA ~\cite{wu2022eda}, 3D-STMN~\cite{2308.16632} and our model on the ScanRefer dataset. As illustrated in (a-c), our method has a superior perception of the referring expression. In the case of (b), our method demonstrates a remarkable capability to accurately identify the object indicated by the text, even within extremely complex text representations. In contrast, 3D-STMN~\cite{2308.16632} and EDA ~\cite{wu2022eda} fail to achieve the same level of accuracy in similar challenging scenarios, for 3D-STMN~\cite{2308.16632}, although it correctly identified some points, it largely mispredicted the mask. Furthermore, in our MCLN, the predicted bounding boxes and masks exhibit a high degree of consistency, indicating the effectiveness of our model.

\section{Conclusion}
We propose MCLN, a novel Multi-branch Collaborative Learning Network designed for the 3D Visual Grounding task, which consists of the 3DREC and 3DRES tasks. Specifically, to make the model better learn the specific information of the two tasks and avoid one being the vassal of the other, we designed two branches to learn the corresponding tasks. Furthermore, to enhance the collaborative training of the two branches and harness their complementary strengths, we introduced the Relative Superpoint Aggregation (RSA) module and the Adaptive Soft Alignment (ASA) module. Extensive experiments and ablation studies demonstrate the effectiveness of our carefully designed modules and the superiority of our model, proving its ability to harmoniously unify 3DREC and 3DRES tasks. 
\clearpage  % TODO REVIEW/FINAL: This \clearpage needs to be removed from both review and camera-ready versions.
\newpage
\subsubsection{Acknowledgements.}
This work was supported by National Science and Technology Major Project (No. 2022ZD0118201), the National Science Fund for Distinguished Young Scholars (No.62025603), the National Natural Science Foundation of China (No. U21B2037, No. U22B2051, No. 62072389), the National Natural Science Fund for Young Scholars of China (No. 62302411), China Postdoctoral Science Foundation (No. 2023M732948), the Natural Science Foundation of Fujian Province of China (No.2021J01002,  No.2022J06001), and partially sponsored by CCF-NetEase ThunderFire Innovation Research Funding (NO. CCF-Netease 202301).

% ---- Bibliography ----
%
% BibTeX users should specify bibliography style 'splncs04'.
% References will then be sorted and formatted in the correct style.
%
\bibliographystyle{splncs04}
\bibliography{main}
\newpage
\appendix
\renewcommand{\thesection}{\Alph{section}}
\section{Appendix}
\subsection{More Performance Comparisons}
Furthermore, we assess the efficacy of our method on the SR3D and NR3D datasets. Notably, our model is the pioneering approach to quantify the performance of the 3DRES task specifically on the SR3D and NR3D datasets. We also evaluate our method's performance the the task proposed by EDA\cite{wu2022eda}: Grounding without Object Name (VG-w/o-ON).
\subsubsection{SR3D/NR3D.} 
SR3D/NR3D is derived from ScanNet, encompassing 83,572 automated, straightforward machine-generated descriptions in the case of SR3D and 41,503 descriptions in NR3D, resembling the human annotations employed in ScanRefer. The primary difference lies in the ScanRefer setup, where the task entails both object detection and matching. Conversely, SR3D/NR3D does not require this dual process.
Tab.~\ref{tab:SR3D_result} presents the accuracy results on the SR3D and NR3D datasets, demonstrating our outstanding performance with $68.4\%$ and $59.8\%$ of Acc@0.25 in the 3DREC task. Especially in the NR3D dataset, our model overperforms the 3DRefTR~\cite{lin2023unified} by a large margin of \textbf{7.2\%}, proving the effectiveness of our method. 
In particular, as depicted in Table~\ref{sr3dnr3d}, our approach consistently demonstrates exceptional performance in 3DRES tasks, achieving noteworthy mIOU of \textbf{49.84\%} and \textbf{46.09\%} in SR3D/NR3D dataset. These results underscore the universality of our method, emphasizing its superior predictive capabilities across both 3DREC and 3DRES tasks.

\subsubsection{Grounding without Object Name (VG-w/o-ON).}
Following EDA\cite{wu2022eda}, we also evaluate our method on the task: grounding objects without mentioning object names (VG-w/o-ON). Specifically, the task needs to replace the object's name with"object" in the ScanRefer validation set. For instance: "This is a brown wooden chair" becomes "This is a brown wooden object". As illustrated in Table~\ref{scanrefer_wo_name}, our model demonstrates remarkable performance enhancements compared to EDA~\cite{wu2022eda}, with a notable increase of \textbf{8.89\%} and \textbf{5.63\%} in Acc@0.25 and Acc@0.5 respectively in the 3DREC task, proving the effectiveness of our approach in spatial relationship perception. Additionally, our model consistently achieves commendable performance in the 3DRES task, underscoring the generality and consistency of our method.

\input{table/Nr3D_Sr3D_buchong}
\input{table/scanrefer_wo_name_buchong}

\subsection{More Ablation Studys}

\input{table/sr3dnr3d_focaldice_buchong}

\begin{figure}[!t] % 使用figure*使图片横跨双栏宽度
  \centering 
  \includegraphics[width=\columnwidth]{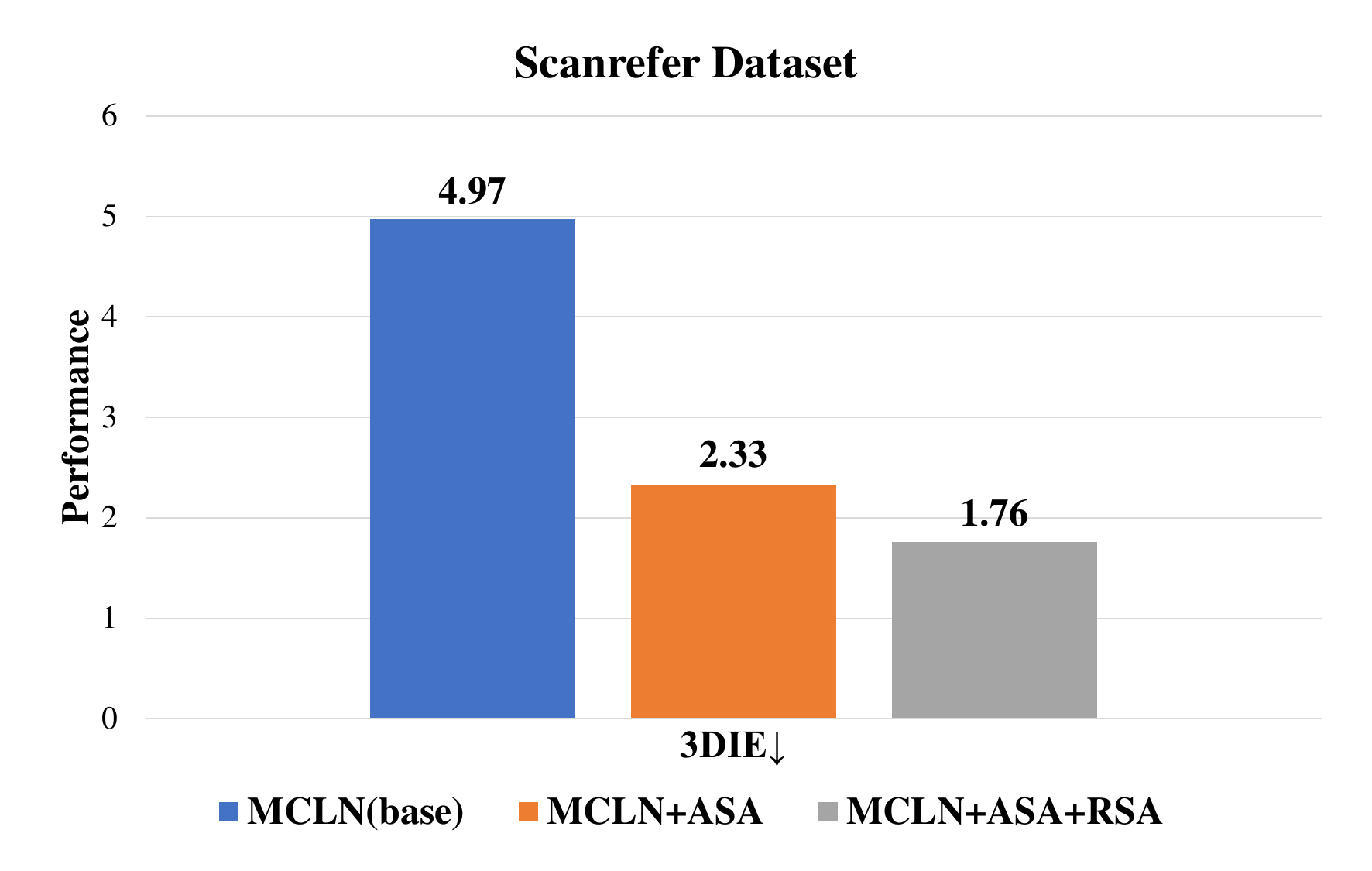} % 使用\textwidth确保图片横跨双栏宽度
  \caption{Ablation study on the 3DIE metric, our ASA module effectively decreases the 3DIE value, proving its effectiveness in alignment.}
  % \vspace{-1em}
  \label{fig:zhuzhuangtu}
\end{figure}
We performed more detailed ablation experiments, comparing our well-designed adaptive focal loss and dice loss with standard focal and dice losses. Additionally, we introduce a novel metric, \textbf{3DIE}, to assess the coherence between 3DREC and 3DRES predictions. This metric serves to validate the effectiveness of each module in our method, shedding light on their contributions to enhancing prediction consistency.
\subsubsection{Ablation study of Adaptive weighted losses.} 
In Tab.~\ref{normal_loss}, the “norm” means the ASA module replaces the adaptive losses with the standard Focal loss and Dice loss. As demonstrated in Tab.~\ref{normal_loss}, our Adaptive Soft Alignment (ASA) module, enhanced with adaptive losses, outperforms the module using standard losses across all metrics. This substantiates that our meticulously crafted adaptive weighted Focal loss and Dice loss are more effective in capturing the distinctions between the outputs of the two branches. Consequently, this facilitates more efficient coordination and alignment, proving our method's effectiveness.

\subsubsection{Ablation study of the 3DIE metric.} 

Inspired by \cite{Luo_Zhou_Sun_Cao_Wu_Deng_Ji_2020}, we introduce a novel metric termed 3D Inconsistency Error (3DIE) to quantitatively assess the degree of prediction conflict between the outputs of the 3DREC and 3DRES branches. Specifically, we define a prediction conflict when one of the branch outputs achieves an intersection over union (IoU) with the ground truth above 0.5, while the other falls below 0.25. Illustrated in Fig.~\ref{fig:zhuzhuangtu}, we did an ablation study on the Scanrefer dataset to validate the effectiveness of our proposed two modules in the 3DIE metric. The Adaptive Soft Alignment (ASA) module and the Relative Superpoint Aggregation (RSA) module consistently exhibit improvements in mitigating prediction conflicts. Notably, the ASA module, in particular, reduces conflicts to less than half (\textbf{4.97\% vs. 2.33\%}), underscoring its efficacy in aligning predictions from the two branches. Additionally, our model exhibits only a \textbf{1.76\%} occurrence of cases with prediction contradiction, further underscoring the superior performance of our approach.

\begin{figure*}[!h]
\centering 
\includegraphics[width=1\columnwidth]{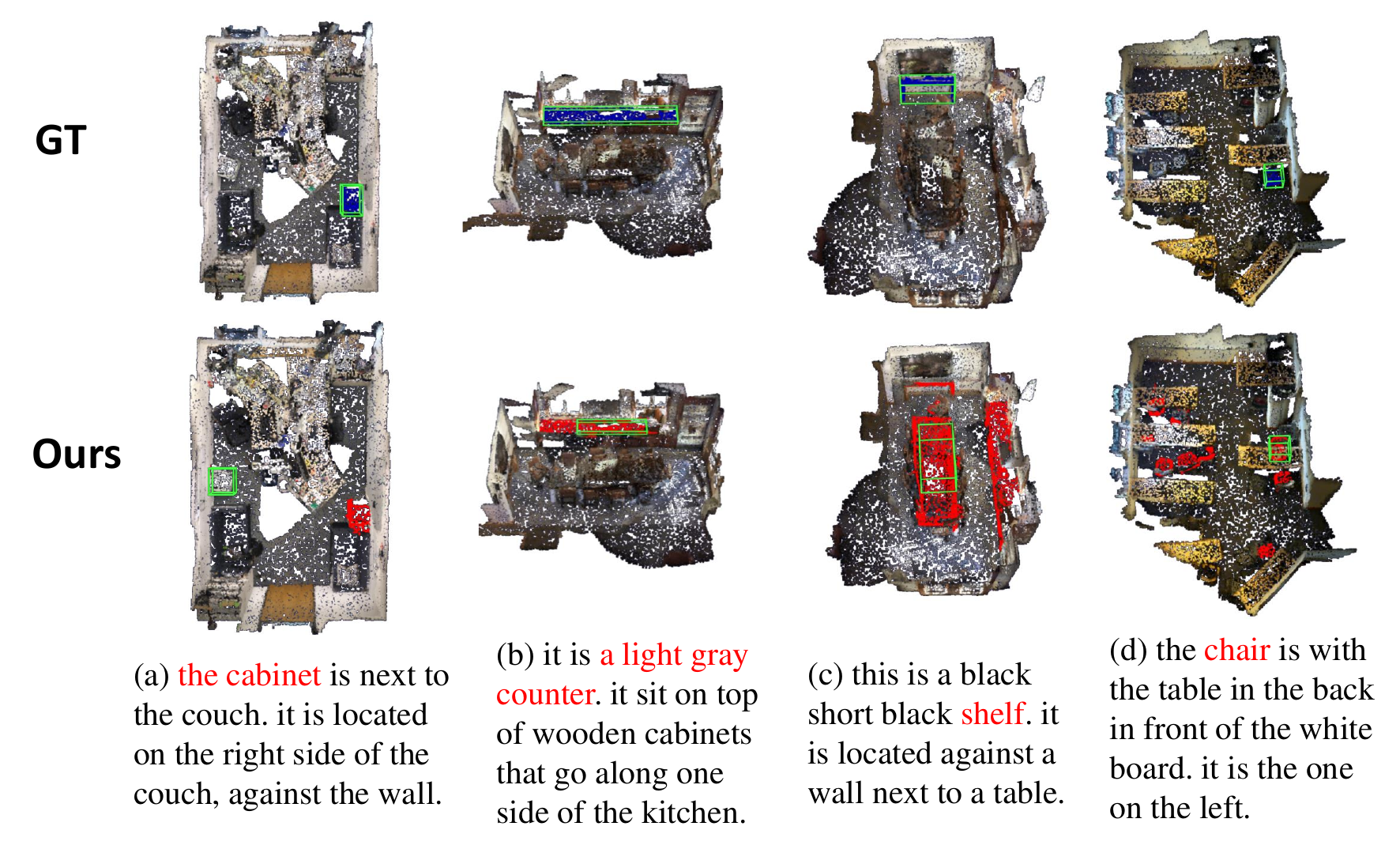}
  % \vspace{-2em}
  \caption{Failure cases:(a) and (b) show the inconsistency between predicted masks and bounding boxes, and (c) and (d) showcase cases where the model generates discrete masks alongside inaccuracies in the predicted bounding boxes.}
  % \vspace{-1em}
  \label{fig:failure_cases}
  % \vspace{-1em}
\end{figure*}

\subsection{More Visualization}
\subsubsection{Failure case analysis.} 
While our method has achieved state-of-the-art performance, it is essential to acknowledge that a noteworthy number of failure instances persist. These occurrences have been subject to thorough visual analysis, as illustrated in Fig.~\ref{fig:failure_cases}. In instances (a) and (b), our model accurately predicts masks; however, there are inaccuracies in the predicted bounding box positions or sizes. In cases (c) and (d), the model exhibits erroneous predictions of masks, with discrete superpoints, coupled with inaccuracies in the predicted bounding boxes. These failure cases underscore the existing limitations in our model's ability to predict masks and bounding boxes. In the realm of masks, there is a potential for predicting redundant superpoints, while in the case of bounding boxes, there may be inaccuracies in predicting their positions and sizes. These limitations not only impact the individual tasks but may also introduce challenges in achieving proper alignment between the predicted outputs.

\end{document}

%% file: table/REC_table.tex
\begin{table*}[t]
\centering
\vspace{-1em}
\caption{The 3D visual grounding results on ScanRefer, accuracy evaluated by IoU 0.25 and IoU 0.5. As depicted in the table, our method attains state-of-the-art (SOTA) performance in both two-stage and single-stage methods. }
% Notably, in two-stage methods, our approach outperforms others in all performance metrics by a large margin.} We color code \colorh{best}, \colorm{second best} and \colorl{third best} performances.}
% \vspace{-1em}
\setlength{\tabcolsep}{0pt}
\resizebox{1\textwidth}{!}{
\begin{tabular}{lcc|cc|cc|cc}
\toprule
\rowcolor[HTML]{E6E6E6}            &  &  & \multicolumn{2}{c|}{Unique ($\sim$19\%)} & \multicolumn{2}{c|}{Multiple ($\sim$81\%)} & \multicolumn{2}{c}{\textbf{Overall}}            \\
\rowcolor[HTML]{E6E6E6} \multirow{-2}{*}{Method}                       &  \multirow{-2}{*}{Venue}        &      \multirow{-2}{*}{Modality}                     & 0.25              & 0.5              & 0.25               & 0.5               & \textbf{0.25}          & \textbf{0.5}           \\ 
\midrule
\multicolumn{9}{l}{\textbf{\textit{Two-Stage Model} }} \\
\midrule
\multirow{2}{*}{ScanRefer~\cite{chen2020scanrefer}}        & \multirow{2}{*}{ECCV2020} & 3D                        & 67.64             & 46.19            & 32.06              & 21.26             & 38.97                  & 26.10                   \\
                                  &                                                            & 3D+2D                     & 76.33             & 53.51            & 32.73              & 21.11             & 41.19                  & 27.40                   \\ 
 ReferIt3D~\cite{achlioptas2020referit3d}                         & ECCV2020            & 3D                        & 53.8              & 37.5             & 21.0               & 12.8              & 26.4                   & 16.9                   \\ 
TGNN~\cite{Huang_Lee_Chen_Liu_2021}                              & AAAI2021                      & 3D                        & 68.61             & 56.80             & 29.84              & 23.18             & 37.37                  & 29.70                   \\ 
 InstanceRefer~\cite{yuan2021instancerefer}                     & ICCV2021              & 3D                        & 77.45             & 66.83            & 31.27              & 24.77             & 40.23                  & 32.93                  \\ 
SAT~\cite{yang2021sat}                               & ICCV2021                        & 3D+2D                     & 73.21             & 50.83            & 37.64              & 25.16             & 44.54                  & 30.14                  \\ 
FFL-3DOG~\cite{feng2021free}                          & ICCV2021                       & 3D                        & 78.80              &  {67.94}       & 35.19              & 25.70              & 41.33                  & 34.01                  \\ 
\multirow{2}{*}{3DVG-Trans~\cite{zhao20213dvg}} & \multirow{2}{*}{ICCV2021}      & 3D                        & 77.16             & 58.47            & 38.38              & 28.70              & 45.90                   & 34.47                  \\
                                  &                                                            & 3D+2D                     & 81.93             & 60.64            & 39.30               & 28.42             & 47.57                  & 34.67                  \\ 
 3D-SPS~\cite{luo20223d}           & CVPR2022               & 3D+2D                      &{84.12}       &  66.72       & 40.32              & 29.82             & 48.82                  & 36.98                  \\ 
\multirow{2}{*}{3DJCG~\cite{cai20223djcg}}            & \multirow{2}{*}{CVPR2022}      & 3D                        & 78.75             & 61.30             & 40.13              & 30.08             & 47.62                  & 36.14                  \\
                                  &                                                            & 3D+2D                     & 83.47        & 64.34            &  41.39         & 30.82        &  49.56             & 37.33             \\ 
 BUTD-DETR~\cite{jain2022bottom}                       & ECCV2022                    & 3D                        & 82.88              & 64.98             & 44.73         & 33.97        & 50.42             & {38.60}             \\ 
D3Net~\cite{chen2021d3net}                        & ECCV2022                    & 3D+2D                        & -              & {{70.35}}             & -         & 30.50        & -             & {37.87}             \\ 
 {EDA}~\cite{wu2022eda}                      & CVPR2023                                                          & 3D                        & {85.76}    & {68.57}   & {49.13}     & {37.64}   & {54.59 } & {42.26} \\ 
3DRefTR~\cite{lin2023unified}             & Arxiv                                                             & 3D                        & {86.12}    & {71.04}   & {50.07}     & {38.65}   & {55.45 } & {43.48} \\ 
MCLN                              & -                         & 3D                       &\colorh{\textbf{86.89}}             &\colorh{\textbf{72.73}}          & \colorh{\textbf{51.96}}             & \colorh{\textbf{40.76}}          & \colorh{\textbf{57.17}}           &\colorh{\textbf{45.53}}                  \\ 
\midrule
\multicolumn{9}{l}{\textbf{\textit{Single-Stage Model} }} \\
% \textbf{\textit{single-stage mode: } }        &         &                        &              &           &             &            &               &                   \\
\midrule
3D-SPS~\cite{luo20223d}           & CVPR2022         & 3D                        & 81.63             & 64.77            & 39.48              & 29.61             & 47.65                  & 36.43                  \\
BUTD-DETR~\cite{jain2022bottom}                    & ECCV2022                                                          & 3D                        & 81.47    & 61.24   & 44.20     & 32.81    & 49.76 & 37.05 \\
{EDA }~\cite{wu2022eda}                      & CVPR2023                                                          & 3D                        & \textbf{86.40}    & \textbf{69.42}   & {49.03}     & {37.93}    & {53.83} & {41.70} \\

3DRefTR~\cite{lin2023unified}             & Arxiv                  & 3D                        &86.40        & 68.01    & {48.82}        &{37.83}      & \textbf{54.43}           &{42.33}  \\ 
MCLN               & -                  & 3D                        &84.43        & 68.36     & \textbf{49.72}        &\textbf{38.41}      & {54.30}           &\textbf{42.64}  \\ 
\bottomrule
\end{tabular}}

\label{tab:REC}
% \vspace{-2em}
\end{table*}

%% file: table/RES_table.tex
\begin{table*}[t]
\small
\centering
% \vspace{-1em}
\caption{The 3D referring expression segmentation results on ScanRefer. In comparison to the previous methods, our method outperforms existing state-of-the-art works by a substantial margin. Notably, it achieves over 5\% increase in mIOU.}
\setlength{\tabcolsep}{0pt}
% \vspace{-1em}

% \renewcommand{\arraystretch}{1.0}
% \setlength{\abovecaptionskip}{2pt}
\resizebox{1\textwidth}{!}{
\begin{tabular}{ccc|cc|cc|cc|c}
\toprule
\rowcolor[HTML]{E6E6E6}            &  &  & \multicolumn{2}{c|}{Unique ($\sim$19\%)} & \multicolumn{2}{c|}{Multiple ($\sim$81\%)} & \multicolumn{2}{c|}{\textbf{Overall}}      &    \\
\rowcolor[HTML]{E6E6E6} \multirow{-2}{*}{Method}                       &  \multirow{-2}{*}{Venue}        &      \multirow{-2}{*}{Modality}                     & 0.25              & 0.5              & 0.25               & 0.5               & \textbf{0.25}          & \textbf{0.5}      & \multirow{-2}{*}{\textbf{mIOU}}       \\ 
\midrule
TGNN~\cite{Huang_Lee_Chen_Liu_2021}                              & AAAI2021                      & 3D                        & -             & -             & -              & -             & 37.50                  & 31.40         & 27.80          \\ 
X-RefSeg3D~\cite{qian2024x}                              & AAAI2024                      & 3D                        & -             & -             & -              & -             & 40.33                  & 33.77         & 29.94          \\ 
3D-STMN ~\cite{2308.16632}               & AAAI2024                     & 3D                        & 89.30            & \colorh{\textbf{84.00}}             & 46.20              & 29.20             & 54.60                  & 39.80            & 39.50       \\ 
3DRefTR~\cite{lin2023unified}                              & Arxiv                     & 3D                        & 87.88             & 69.77             & 51.61              & 41.91            & 57.02                  & 46.07         & 40.76          \\ 
MCLN                              & -                         & 3D                        &\colorh{\textbf{89.57}}              & 78.22        & \colorh{\textbf{53.28}}           & \colorh{\textbf{45.88}}           & \colorh{\textbf{58.70}}       & \colorh{\textbf{50.70}}     &\colorh{\textbf{44.72}}  \\ 

\bottomrule
\end{tabular}}
 %We color code \colorh{best}, \colorm{second best} and \colorl{third best} performances.}
% \vspace{-1em}
\label{tab:RES}
\end{table*}

%% file: table/Alignment_and_mulplus.tex
\begin{table*}[t]
    \begin{minipage}{0.48\textwidth}
        \caption{Ablation study to assess the influence of alignment targets using either bounding boxes or query masks.}
        % \vspace{-1em}
        \begin{tabular}{c|cc|ccl}
            \toprule
            \rowcolor[HTML]{E6E6E6} & \multicolumn{2}{c|}{\textbf{3DREC}} & \multicolumn{3}{c}{\textbf{3DRES}} \\ 
            \rowcolor[HTML]{E6E6E6} \multirow{-2}{*}{Aligns} & \textbf{0.25} & \multicolumn{1}{c|}{\textbf{0.5}} & \textbf{0.25} & \textbf{0.5} & \textbf{mIOU} \\ \midrule
            Box & 55.62 & 44.23 & 55.77 & 49.17 & 43.41 \\
            Mask & \textbf{57.17} & \textbf{45.53} & \textbf{58.70} & \textbf{50.70} & \textbf{44.72} \\ \midrule
        \end{tabular}
        \label{Alignment}
        % \vspace{-2em}
    \end{minipage}%
    \hfill
    \begin{minipage}{0.48\textwidth}
        \caption{Ablation study to assess the effectiveness of different feature fusion methods in the RSA module.}
        % \vspace{-1em}
        \begin{tabular}{c|cc|ccl}
            \toprule
            \rowcolor[HTML]{E6E6E6} & \multicolumn{2}{c|}{\textbf{3DREC}} & \multicolumn{3}{c}{\textbf{3DRES}} \\ 
            \rowcolor[HTML]{E6E6E6} \multirow{-2}{*}{Fusion} & \textbf{0.25} & \multicolumn{1}{c|}{\textbf{0.5}} & \textbf{0.25} & \textbf{0.5} & \textbf{mIOU} \\ \midrule
            Mul & 55.79 & 44.22 & 57.29 & 50.62 & 44.64 \\
            Plus & \textbf{57.17} & \textbf{45.53} & \textbf{58.70} & \textbf{50.70} & \textbf{44.72} \\ \midrule
        \end{tabular}
        \label{Fusion}
        % \vspace{-2em}
    \end{minipage}
\end{table*}

%% file: table/RSA-ASA_and_Focal-Dice.tex
\begin{table*}[t]
\begin{minipage}[b]{0.48\linewidth}
    
    \caption{Ablation study on ASA module and RSA module. We add the ASA module first, then the RSA module.}
        % \vspace{-1em}
    \resizebox{1\textwidth}{!}{
    \begin{tabular}{l|cc|ccl}
        \toprule
        \rowcolor[HTML]{E6E6E6} & \multicolumn{2}{c|}{\textbf{3DREC}} & \multicolumn{3}{c}{\textbf{3DRES}} \\ 
        \rowcolor[HTML]{E6E6E6} \multirow{-2}{*}{Modules} & \textbf{0.25} & \multicolumn{1}{c|}{\textbf{0.5}} & \textbf{0.25} & \textbf{0.5} & \textbf{mIOU} \\ \midrule
        Base & 55.01 & 42.39 & 58.59 & 45.28 & 41.18 \\
        {+ASA} & 56.75 & 45.29 & 58.15 & 48.20 & 42.85 \\
        {+ASA+RSA} & \textbf{57.17} & \textbf{45.53} & \textbf{58.70} & \textbf{50.70} & \textbf{44.72} \\ \midrule
    \end{tabular}}
    \label{ASA}
    % \vspace{-2em}
\end{minipage}%
\hfill
\begin{minipage}[b]{0.48\linewidth}
    \caption{Ablation study on the point-wise Focal loss and mask-wise Dice loss.}
        % \vspace{-1em}
    \resizebox{1\textwidth}{!}{
    \begin{tabular}{cc|cc|ccl}
        \toprule
        \rowcolor[HTML]{E6E6E6}  & & \multicolumn{2}{c|}{\textbf{3DREC}} & \multicolumn{3}{c}{\textbf{3DRES}} \\ 
        \rowcolor[HTML]{E6E6E6} \multirow{-2}{*}{Focal} & \multirow{-2}{*}{Dice} & \textbf{0.25} & \multicolumn{1}{c|}{\textbf{0.5}} & \textbf{0.25} & \textbf{0.5} & \textbf{mIOU} \\ \midrule
        - & - & 55.01 & 42.39 & 58.59 & 45.28 & 41.18 \\
        point & - & 56.02 & 45.02 & 57.73 & 50.13 & 44.03 \\
        - & mask & {56.34} & {44.76} & {58.04} & {50.26} & {44.31} \\
        point & mask & \textbf{57.17} & \textbf{45.53} & \textbf{58.70} & \textbf{50.70} & \textbf{44.72} \\ \midrule
    \end{tabular}}
    \label{Ablation_Point_Wise}
    % \vspace{-2em}
\end{minipage}
\end{table*}

%% file: table/Nr3D_Sr3D_buchong.tex
\begin{table}[!t]
\small
\centering
\setlength{\tabcolsep}{4.5pt}
\setlength{\abovecaptionskip}{2pt}
\caption{For the 3DREC task on SR3D and NR3D datasets, we assess performance using Acc@0.25IoU as the metric.}
\begin{tabular}{ccc|cc}
\toprule
\rowcolor[HTML]{E6E6E6} Method                        & Venue        & Modality                            & \textbf{SR3D}          & \textbf{NR3D}         \\
\midrule
ReferIt3D \footnotesize{\cite{achlioptas2020referit3d}}     & ECCV20      & 3D                  &39.8       & 35.6        \\
TGNN \footnotesize{\cite{Huang_Lee_Chen_Liu_2021}}        & AAAI21      &3D                                 &45.0            & 37.3          \\
TransRefer3D \footnotesize{\cite{He_Zhao_Luo_Hui_Huang_Zhang_Liu_2021}}       &MM21     &3D                    &  57.4          &  42.1         \\
InstanceRefer \footnotesize{\cite{yuan2021instancerefer}}       &ICCV21       &3D                & 48.0           &  38.8        \\
3DVG-Transfor. \footnotesize{\cite{zhao20213dvg}}     &ICCV21       &3D                      & 51.4         & 40.8         \\
FFL-3DOG \footnotesize{\cite{feng2021free}}     &ICCV21       &3D                              & -           &  41.7           \\
SAT \footnotesize{\cite{yang2021sat}}       &ICCV21       &3D+2D                                    & 57.9        & 49.2     \\
3DReferTrans. \footnotesize{\cite{Abdelreheem_Upadhyay_Skorokhodov_Yahya_Chen_Elhoseiny}}     &WACV22     &3D & 47.0        &  39.0           \\
LanguageRefer \footnotesize{\cite{Roh_Desingh_Farhadi_Fox_2021}}        &CoRL22       &3D                 & 56.0       &  43.9           \\
3D-SPS~\footnotesize{\cite{luo20223d}}        &CVPR22       &3D+2D                 & 62.6           & {51.5}          \\
BUTD-DETR  \footnotesize{\cite{jain2022bottom}}       &ECCV22       &3D                        &  65.6  &  {49.1}         \\
LAR~\footnotesize{\cite{Bakr_Alsaedy_Elhoseiny_2022}}        &NeurIPS22     &3D                        &  59.6    &   48.9        \\
{EDA}~\footnotesize{\cite{wu2022eda}}        &CVPR2023     &3D                                                & 68.1 &  {52.1}\\ 
{3DRefTR}~\footnotesize{\cite{lin2023unified}}        &Arxiv    &3D                                                & \colorh{\textbf{68.5}} &  {52.6}\\ 

MCLN      & -     &3D                                                             &68.4 & \colorh{\textbf{59.8}} \\ 
\bottomrule
% \vspace{-0.5em}
\end{tabular}
% \vspace{-0.5em}
\label{tab:SR3D_result}
\end{table}

%% file: table/scanrefer_wo_name_buchong.tex
\begin{table}[t]
        \caption{Performance of visual grounding without object name.}
        \centering
        \begin{tabular}{c|cc|ccl}
            \toprule
            \rowcolor[HTML]{E6E6E6} & \multicolumn{2}{c|}{\textbf{3DREC}} & \multicolumn{3}{c}{\textbf{3DRES}} \\ 
            \rowcolor[HTML]{E6E6E6} \multirow{-2}{*}{Method} & \textbf{0.25} & \multicolumn{1}{c|}{\textbf{0.5}} & \textbf{0.25} & \textbf{0.5} & \textbf{mIOU} \\ \midrule
            Scanrefer\cite{chen2020scanrefer} & 10.51 & 6.20 & - & - & - \\
            TGNN\cite{Huang_Lee_Chen_Liu_2021} & {11.64} & {9.51} & {-} & {-} & {-}\\
            Instancerefer\cite{yuan2021instancerefer} & 13.92 & 11.47 & - & - & - \\
            BUTD-DETR\cite{jain2022bottom} & 11.99 & 8.95 & - & - & - \\
            EDA\cite{wu2022eda} & 26.50 & 21.20 & - & - & - \\
            MCLN & \colorh{\textbf{35.39}} & \colorh{\textbf{26.83}} & \colorh{\textbf{36.66}}& \colorh{\textbf{30.96}} & \colorh{\textbf{27.41}} \\
            \bottomrule
        \end{tabular}
        % \vspace{-1em}
        \label{scanrefer_wo_name}
        % \vspace{-1em}
\end{table}

%% file: table/sr3dnr3d_focaldice_buchong.tex
\begin{table*}[!t]
    \begin{minipage}{0.48\textwidth}
        \caption{Performance of our model on SR3D/NR3D datasets.}
        % \vspace{-1em}
        \begin{tabular}{c|cc|ccl}
            \toprule
            \rowcolor[HTML]{E6E6E6} & \multicolumn{2}{c|}{\textbf{3DREC}} & \multicolumn{3}{c}{\textbf{3DRES}} \\ 
            \rowcolor[HTML]{E6E6E6} \multirow{-2}{*}{Dataset} & \textbf{0.25} & \multicolumn{1}{c|}{\textbf{0.5}} & \textbf{0.25} & \textbf{0.5} & \textbf{mIOU} \\ \midrule
            SR3D & 68.43 & 57.30 & 64.70 & 57.88 & 49.84 \\
            NR3D & {59.82} & {51.38} & {58.35} & {52.68} & {46.09} \\ \midrule
        \end{tabular}
        % \vspace{-1em}
        \label{sr3dnr3d}
        % \vspace{-1.5em}
    \end{minipage}%
    \hfill
    \begin{minipage}{0.48\textwidth}
        \caption{Ablation study on our losses and normal losses.}
        % \vspace{-1em}
        \begin{tabular}{cc|cc|ccl}
            \toprule
            \rowcolor[HTML]{E6E6E6}  & & \multicolumn{2}{c|}{\textbf{3DREC}} & \multicolumn{3}{c}{\textbf{3DRES}} \\ 
            \rowcolor[HTML]{E6E6E6} \multirow{-2}{*}{Focal} & \multirow{-2}{*}{Dice} & \textbf{0.25} & \multicolumn{1}{c|}{\textbf{0.5}} & \textbf{0.25} & \textbf{0.5} & \textbf{mIOU} \\ \midrule
            norm & norm & 56.05 & 44.13 & 57.69 & 50.22 & 44.24 \\
            point & mask & \textbf{57.17} & \textbf{45.53} & \textbf{58.70} & \textbf{50.70} & \textbf{44.72} \\ \midrule
        \end{tabular}
        % \vspace{-1em}
        \label{normal_loss}
        % \vspace{-1.5em}
    \end{minipage}
\end{table*}

%% file: main.bbl
\begin{thebibliography}{10}
\providecommand{\url}[1]{\texttt{#1}}
\providecommand{\urlprefix}{URL }
\providecommand{\doi}[1]{https://doi.org/#1}

\bibitem{Abdelreheem_Upadhyay_Skorokhodov_Yahya_Chen_Elhoseiny}
Abdelreheem, A., Upadhyay, U., Skorokhodov, I., Yahya, R., Chen, J., Elhoseiny, M.: 3dreftransformer: Fine-grained object identification in real-world scenes using natural language

\bibitem{achlioptas2020referit3d}
Achlioptas, P., Abdelreheem, A., Xia, F., Elhoseiny, M., Guibas, L.: Referit3d: Neural listeners for fine-grained 3d object identification in real-world scenes. In: Computer Vision--ECCV 2020: 16th European Conference, Glasgow, UK, August 23--28, 2020, Proceedings, Part I 16. pp. 422--440. Springer (2020)

\bibitem{Bakr_Alsaedy_Elhoseiny_2022}
Bakr, E., Alsaedy, Y., Elhoseiny, M.: Look around and refer: 2d synthetic semantics knowledge distillation for 3d visual grounding  (Nov 2022)

\bibitem{Bolya_Zhou_Xiao_Lee_2019}
Bolya, D., Zhou, C., Xiao, F., Lee, Y.J.: Yolact: Real-time instance segmentation. In: 2019 IEEE/CVF International Conference on Computer Vision (ICCV) (Oct 2019). \doi{10.1109/iccv.2019.00925}, \url{http://dx.doi.org/10.1109/iccv.2019.00925}

\bibitem{cai20223djcg}
Cai, D., Zhao, L., Zhang, J., Sheng, L., Xu, D.: 3djcg: A unified framework for joint dense captioning and visual grounding on 3d point clouds. In: Proceedings of the IEEE/CVF Conference on Computer Vision and Pattern Recognition. pp. 16464--16473 (2022)

\bibitem{Caruana_Pratt_Thrun_2017}
Caruana, R., Pratt, L., Thrun, S.: Multitask Learning *, p. 893–893 (Jan 2017). \doi{10.1007/978-1-4899-7687-1_100322}, \url{http://dx.doi.org/10.1007/978-1-4899-7687-1_100322}

\bibitem{chen2020scanrefer}
Chen, D.Z., Chang, A.X., Nie{\ss}ner, M.: Scanrefer: 3d object localization in rgb-d scans using natural language. In: European conference on computer vision. pp. 202--221. Springer (2020)

\bibitem{chen2021d3net}
Chen, D.Z., Wu, Q., Nie{\ss}ner, M., Chang, A.X.: D3net: a speaker-listener architecture for semi-supervised dense captioning and visual grounding in rgb-d scans  (2021)

\bibitem{chen2022language}
Chen, S., Guhur, P.L., Tapaswi, M., Schmid, C., Laptev, I.: Language conditioned spatial relation reasoning for 3d object grounding. Advances in neural information processing systems  \textbf{35},  20522--20535 (2022)

\bibitem{Dai_Chang_Savva_Halber_Funkhouser_Niessner_2017}
Dai, A., Chang, A.X., Savva, M., Halber, M., Funkhouser, T., Niessner, M.: Scannet: Richly-annotated 3d reconstructions of indoor scenes. In: 2017 IEEE Conference on Computer Vision and Pattern Recognition (CVPR) (Jul 2017). \doi{10.1109/cvpr.2017.261}, \url{http://dx.doi.org/10.1109/cvpr.2017.261}

\bibitem{feng2021free}
Feng, M., Li, Z., Li, Q., Zhang, L., Zhang, X., Zhu, G., Zhang, H., Wang, Y., Mian, A.: Free-form description guided 3d visual graph network for object grounding in point cloud. In: Proceedings of the IEEE/CVF International Conference on Computer Vision. pp. 3722--3731 (2021)

\bibitem{Fu_Shvets_Berg_2019}
Fu, C.Y., Shvets, M., Berg, A.: Retinamask: Learning to predict masks improves state-of-the-art single-shot detection for free. arXiv: Computer Vision and Pattern Recognition,arXiv: Computer Vision and Pattern Recognition  (Jan 2019)

\bibitem{han2020occuseg}
Han, L., Zheng, T., Xu, L., Fang, L.: Occuseg: Occupancy-aware 3d instance segmentation. In: Proceedings of the IEEE/CVF conference on computer vision and pattern recognition. pp. 2940--2949 (2020)

\bibitem{He_Zhao_Luo_Hui_Huang_Zhang_Liu_2021}
He, D., Zhao, Y., Luo, J., Hui, T., Huang, S., Zhang, A., Liu, S.: Transrefer3d: Entity-and-relation aware transformer for fine-grained 3d visual grounding. In: Proceedings of the 29th ACM International Conference on Multimedia (Oct 2021). \doi{10.1145/3474085.3475397}, \url{http://dx.doi.org/10.1145/3474085.3475397}

\bibitem{He_Gkioxari_Dollar_Girshick_2020}
He, K., Gkioxari, G., Dollar, P., Girshick, R.: Mask r-cnn. IEEE Transactions on Pattern Analysis and Machine Intelligence p. 386–397 (Feb 2020). \doi{10.1109/tpami.2018.2844175}, \url{http://dx.doi.org/10.1109/tpami.2018.2844175}

\bibitem{Hua_Liao_Tian_Zhang_Zou_2023}
Hua, G., Liao, M., Tian, S., Zhang, Y., Zou, W.: Multiple relational learning network for joint referring expression comprehension and segmentation. IEEE Transactions on Multimedia p. 1–13 (Jan 2023). \doi{10.1109/tmm.2023.3241802}, \url{http://dx.doi.org/10.1109/tmm.2023.3241802}

\bibitem{Huang_Lee_Chen_Liu_2021}
Huang, P.H., Lee, H.H., Chen, H.T., Liu, T.L.: Text-guided graph neural networks for referring 3d instance segmentation. Proceedings of the ... AAAI Conference on Artificial Intelligence,Proceedings of the ... AAAI Conference on Artificial Intelligence  (May 2021)

\bibitem{jain2022bottom}
Jain, A., Gkanatsios, N., Mediratta, I., Fragkiadaki, K.: Bottom up top down detection transformers for language grounding in images and point clouds. In: European Conference on Computer Vision. pp. 417--433. Springer (2022)

\bibitem{9761944}
Ji, J., Huang, X., Sun, X., Zhou, Y., Luo, G., Cao, L., Liu, J., Shao, L., Ji, R.: Multi-branch distance-sensitive self-attention network for image captioning. IEEE Transactions on Multimedia  \textbf{25},  3962--3974 (2023). \doi{10.1109/TMM.2022.3169061}

\bibitem{9802801}
Ji, J., Ma, Y., Sun, X., Zhou, Y., Wu, Y., Ji, R.: Knowing what to learn: A metric-oriented focal mechanism for image captioning. IEEE Transactions on Image Processing  \textbf{31},  4321--4335 (2022). \doi{10.1109/TIP.2022.3183434}

\bibitem{10.1145/3394171.3414009}
Ji, J., Sun, X., Zhou, Y., Ji, R., Chen, F., Liu, J., Tian, Q.: Attacking image captioning towards accuracy-preserving target words removal. In: Proceedings of the 28th ACM International Conference on Multimedia. p. 4226–4234. MM '20, Association for Computing Machinery, New York, NY, USA (2020). \doi{10.1145/3394171.3414009}, \url{https://doi.org/10.1145/3394171.3414009}

\bibitem{Li_Zhang_Sun_Wu_Zhao_Tan_2022}
Li, Q., Zhang, Y., Sun, S., Wu, J., Zhao, X., Tan, M.: Cross-modality synergy network for referring expression comprehension and segmentation. Neurocomputing p. 99–114 (Jan 2022). \doi{10.1016/j.neucom.2021.09.066}, \url{http://dx.doi.org/10.1016/j.neucom.2021.09.066}

\bibitem{liang2021instance}
Liang, Z., Li, Z., Xu, S., Tan, M., Jia, K.: Instance segmentation in 3d scenes using semantic superpoint tree networks. In: Proceedings of the IEEE/CVF international conference on computer vision. pp. 2783--2792 (2021)

\bibitem{lin2023unified}
Lin, H., Luo, Y., Zheng, X., Li, L., Chao, F., Jin, T., Luo, D., Wang, C., Wang, Y., Cao, L.: A unified framework for 3d point cloud visual grounding. arXiv preprint arXiv:2308.11887  (2023)

\bibitem{Lin_Goyal_Girshick_He_Dollar_2017}
Lin, T.Y., Goyal, P., Girshick, R., He, K., Dollar, P.: Focal loss for dense object detection. In: 2017 IEEE International Conference on Computer Vision (ICCV) (Oct 2017). \doi{10.1109/iccv.2017.324}, \url{http://dx.doi.org/10.1109/iccv.2017.324}

\bibitem{Liu_Ott_Goyal_Du_Joshi_Chen_Levy_Lewis_Zettlemoyer_Stoyanov}
Liu, Y., Ott, M., Goyal, N., Du, J., Joshi, M., Chen, D., Levy, O., Lewis, M., Zettlemoyer, L., Stoyanov, V.: Roberta: A robustly optimized bert pretraining approach

\bibitem{Liu_Zhang_Cao_Hu_Tong_2021}
Liu, Z., Zhang, Z., Cao, Y., Hu, H., Tong, X.: Group-free 3d object detection via transformers. In: 2021 IEEE/CVF International Conference on Computer Vision (ICCV) (Oct 2021). \doi{10.1109/iccv48922.2021.00294}, \url{http://dx.doi.org/10.1109/iccv48922.2021.00294}

\bibitem{Luo_Zhou_Sun_Cao_Wu_Deng_Ji_2020}
Luo, G., Zhou, Y., Sun, X., Cao, L., Wu, C., Deng, C., Ji, R.: Multi-task collaborative network for joint referring expression comprehension and segmentation. In: 2020 IEEE/CVF Conference on Computer Vision and Pattern Recognition (CVPR) (Jun 2020). \doi{10.1109/cvpr42600.2020.01005}, \url{http://dx.doi.org/10.1109/cvpr42600.2020.01005}

\bibitem{luo20223d}
Luo, J., Fu, J., Kong, X., Gao, C., Ren, H., Shen, H., Xia, H., Liu, S.: 3d-sps: Single-stage 3d visual grounding via referred point progressive selection. In: Proceedings of the IEEE/CVF Conference on Computer Vision and Pattern Recognition. pp. 16454--16463 (2022)

\bibitem{ma2023towards}
Ma, Y., Ji, J., Sun, X., Zhou, Y., Ji, R.: Towards local visual modeling for image captioning. Pattern Recognition  \textbf{138},  109420 (2023)

\bibitem{ma2022xclip}
Ma, Y., Xu, G., Sun, X., Yan, M., Zhang, J., Ji, R.: X-clip: End-to-end multi-grained contrastive learning for video-text retrieval. In: Proceedings of the 30th ACM International Conference on Multimedia. pp. 638--647 (2022)

\bibitem{ma2023xmesh}
Ma, Y., Zhang, X., Sun, X., Ji, J., Wang, H., Jiang, G., Zhuang, W., Ji, R.: X-mesh: Towards fast and accurate text-driven 3d stylization via dynamic textual guidance. In: Proceedings of the IEEE/CVF International Conference on Computer Vision. pp. 2749--2760 (2023)

\bibitem{Milletari_Navab_Ahmadi_2016}
Milletari, F., Navab, N., Ahmadi, S.A.: V-net: Fully convolutional neural networks for volumetric medical image segmentation. In: 2016 Fourth International Conference on 3D Vision (3DV) (Oct 2016). \doi{10.1109/3dv.2016.79}, \url{http://dx.doi.org/10.1109/3dv.2016.79}

\bibitem{Nekrasov_Dharmasiri_Spek_Drummond_Shen_Reid_2018}
Nekrasov, V., Dharmasiri, T., Spek, A., Drummond, T., Shen, C., Reid, I.: Real-time joint semantic segmentation and depth estimation using asymmetric annotations. arXiv: Computer Vision and Pattern Recognition,arXiv: Computer Vision and Pattern Recognition  (Sep 2018)

\bibitem{Qi_Yi_Su_Guibas_2017}
Qi, C., Yi, L., Su, H., Guibas, L.: Pointnet++: Deep hierarchical feature learning on point sets in a metric space. Cornell University - arXiv,Cornell University - arXiv  (Jun 2017)

\bibitem{qian2024x}
Qian, Z., Ma, Y., Ji, J., Sun, X.: X-refseg3d: Enhancing referring 3d instance segmentation via structured cross-modal graph neural networks. In: Proceedings of the AAAI Conference on Artificial Intelligence. vol.~38, pp. 4551--4559 (2024)

\bibitem{Rezatofighi_Tsoi_Gwak_Sadeghian_Reid_Savarese_2019}
Rezatofighi, H., Tsoi, N., Gwak, J., Sadeghian, A., Reid, I., Savarese, S.: Generalized intersection over union: A metric and a loss for bounding box regression. In: 2019 IEEE/CVF Conference on Computer Vision and Pattern Recognition (CVPR) (Jun 2019). \doi{10.1109/cvpr.2019.00075}, \url{http://dx.doi.org/10.1109/cvpr.2019.00075}

\bibitem{Roh_Desingh_Farhadi_Fox_2021}
Roh, J., Desingh, K., Farhadi, A., Fox, D.: Languagerefer: Spatial-language model for 3d visual grounding. Cornell University - arXiv,Cornell University - arXiv  (Jul 2021)

\bibitem{Sun_Qing_Tan_Xu_2022}
Sun, J., Qing, C., Tan, J., Xu, X.: Superpoint transformer for 3d scene instance segmentation  (Nov 2022)

\bibitem{2308.16632}
Wu, C., Ma, Y., Chen, Q., Wang, H., Luo, G., Ji, J., Sun, X.: 3d-stmn: Dependency-driven superpoint-text matching network for end-to-end 3d referring expression segmentation (2023)

\bibitem{wu2022eda}
Wu, Y., Cheng, X., Zhang, R., Cheng, Z., Zhang, J.: Eda: Explicit text-decoupling and dense alignment for 3d visual grounding. In: Proceedings of the IEEE Conference on Computer Vision and Pattern Recognition (CVPR) (2023)

\bibitem{yang2024sam}
Yang, D., Ji, J., Ma, Y., Guo, T., Wang, H., Sun, X., Ji, R.: Sam as the guide: Mastering pseudo-label refinement in semi-supervised referring expression segmentation. arXiv preprint arXiv:2406.01451  (2024)

\bibitem{Yang_Ji_Sun_Wang_Li_Ma_Ji_2023}
Yang, D., Ji, J., Sun, X., Wang, H., Li, Y., Ma, Y., Ji, R.: Semi-supervised panoptic narrative grounding. In: Proceedings of the 31st ACM International Conference on Multimedia (Oct 2023). \doi{10.1145/3581783.3612259}, \url{http://dx.doi.org/10.1145/3581783.3612259}

\bibitem{yang2021sat}
Yang, Z., Zhang, S., Wang, L., Luo, J.: Sat: 2d semantics assisted training for 3d visual grounding. In: Proceedings of the IEEE/CVF International Conference on Computer Vision. pp. 1856--1866 (2021)

\bibitem{yuan2021instancerefer}
Yuan, Z., Yan, X., Liao, Y., Zhang, R., Wang, S., Li, Z., Cui, S.: Instancerefer: Cooperative holistic understanding for visual grounding on point clouds through instance multi-level contextual referring. In: Proceedings of the IEEE/CVF International Conference on Computer Vision. pp. 1791--1800 (2021)

\bibitem{zhao20213dvg}
Zhao, L., Cai, D., Sheng, L., Xu, D.: 3dvg-transformer: Relation modeling for visual grounding on point clouds. In: Proceedings of the IEEE/CVF International Conference on Computer Vision. pp. 2928--2937 (2021)

\end{thebibliography}
